\pdfoutput=1

\documentclass[x11names, 11pt]{article}

\usepackage[final]{acl}

\usepackage{times}
\usepackage{latexsym}
\usepackage{fontawesome}
\usepackage{multirow}
\usepackage{booktabs}
\usepackage{boldline} 
\usepackage{rotating}
\usepackage{float}
\usepackage[section]{placeins}
\usepackage[T1]{fontenc}

\usepackage[utf8]{inputenc}

\usepackage{microtype}

\usepackage{inconsolata}

\usepackage{graphicx}

\usepackage{hyperref}
\usepackage{url}
\usepackage{booktabs}
\usepackage{amsfonts}
\usepackage{nicefrac}
\usepackage{microtype}
\usepackage{graphicx}
\usepackage{amssymb}

\usepackage{soul}
\PassOptionsToPackage{x11names}{xcolor}
\usepackage{pgf}

\definecolor{verb-source}{RGB}{0,100,120}
\definecolor{verb-intervened}{RGB}{52, 80,195}
\definecolor{task-source}{RGB}{101,77,77}
\definecolor{task-intervened}{RGB}{1,104,149}
\definecolor{counterfactual}{RGB}{129,43,238}
\definecolor{delectricblue}{RGB}{0,0,205}
\colorlet{lightpurple}{delectricblue!15} 
\definecolor{verbalization}{RGB}{202,225,255}
\colorlet{verbalization}{verbalization!65} 
\definecolor{counterfactual}{RGB}{129,43,238}
\colorlet{counterfactual}{counterfactual!15} 

\usepackage{lipsum}
\usepackage{pifont}
\usepackage{xspace}

\makeatletter
 \def\SOUL@hlpreamble{%
 \setul{}{1.9ex}
 \let\SOUL@stcolor\SOUL@hlcolor
 \SOUL@stpreamble
 }
\newcommand{\hlc}[2][yellow]{{%
    \colorlet{foo}{#1}%
    \sethlcolor{foo}\hl{#2}}%
}

\usepackage{bbm}
\usepackage{amsmath}
\usepackage{amssymb}
\usepackage{mathtools}
\newcommand{\getvals}[3]{{\texttt{GetVals}}(#1,#2, #3)}

\newcommand{\intinv}{\texttt{IntInv}}

\newcommand{\model}{\mathcal{M}}

\newcommand{\inputx}{\mathbf{x}}

\newcommand{\predfunc}{\tau}

\newcommand{\coloredinfer}{\hlc[lightpurple]{\texttt{inference}} function\xspace}
\newcommand{\coloredverb}{\hlc[verbalization]{\texttt{verbalization}} function\xspace}
\newcommand{\coloredcf}{{\hlc[counterfactual]{\texttt{counterfactual output}}}\xspace}
\newcommand{\funcinfer}{{\texttt{inference}} function\xspace}
\newcommand{\funcverb}{{\texttt{verbalization}} function\xspace}
\newcommand{\cf}{{{\texttt{counterfactual} \texttt{output}}}\xspace}

\newcommand{\emldisplay}[2]{\texttt{\href{mailto:#1}{#2}}}

\let\svthefootnote\thefootnote
\newcommand\blankfootnote[1]{%
  \let\thefootnote\relax\footnotetext{#1}%
  \let\thefootnote\svthefootnote%
}

\newif\ifcomments
\commentstrue

\ifcomments
    \providecommand{\jt}[1]{{\protect\color{Magenta3}\fontfamily{lmtt}\selectfont{[JT: #1]}}}
    \providecommand{\nfl}[1]{{\protect\color{Green}\fontfamily{lmtt}\selectfont{[NL: #1]}}}
    \providecommand{\xc}[1]{{\protect\color{ProcessBlue}\fontfamily{lmtt}\selectfont{[XC: #1]}}}
\else
    \providecommand{\jt}[1]{}
    \providecommand{\nfl}[1]{}
    \providecommand{\xc}[1]{}
\fi

\title{
Inference and Verbalization Functions During In-Context Learning
}

\author{
  Junyi Tao\footnotemark[1]\\
  Stanford University \\
  \emldisplay{junyitao@stanford.edu}{junyitao@stanford.edu} \\
  \And
  Xiaoyin Chen\footnotemark[1] \\
  Mila, University of Montreal \\
  \emldisplay{xiaoyin.chen@mila.quebec}{xiaoyin.chen@mila.quebec} \\
  \And
  Nelson F. Liu \\
  Stanford University \\
  \emldisplay{nfliu@cs.stanford.edu}{nfliu@cs.stanford.edu}
}

\begin{document}

\maketitle
\blankfootnote{\llap{\textsuperscript{*}}Equal contribution.}

\begin{abstract}

Large language models (LMs) are capable of in-context learning from a few demonstrations (example-label pairs) to solve new tasks during inference.
Despite the intuitive importance of high-quality demonstrations, previous work has observed that, in some settings, ICL performance is minimally affected by irrelevant labels \citep{min-etal-2022-rethinking}.
We hypothesize that LMs perform ICL with irrelevant labels via two sequential processes: an \funcinfer that solves the task, followed by a \funcverb that maps the inferred answer to the label space. Importantly, we hypothesize that the \funcinfer is invariant to remappings of the label space (e.g., ``true''/``false'' to ``cat''/``dog''), enabling LMs to share the same \funcinfer across settings with different label words. We empirically validate this hypothesis with controlled layer-wise interchange intervention experiments. Our findings confirm the hypotheses on multiple datasets and tasks (natural language inference, sentiment analysis, and topic classification) and further suggest that the two functions can be localized in specific layers across various open-sourced models, including \textsc{Gemma-7B}, \textsc{Mistral-7B-v0.3}, \textsc{Gemma-2-27B}, and \textsc{Llama-3.1-70B}.


\end{abstract}

\section{Introduction}

\begin{figure}[t]
    \centering
    \includegraphics[width=1.0\linewidth]{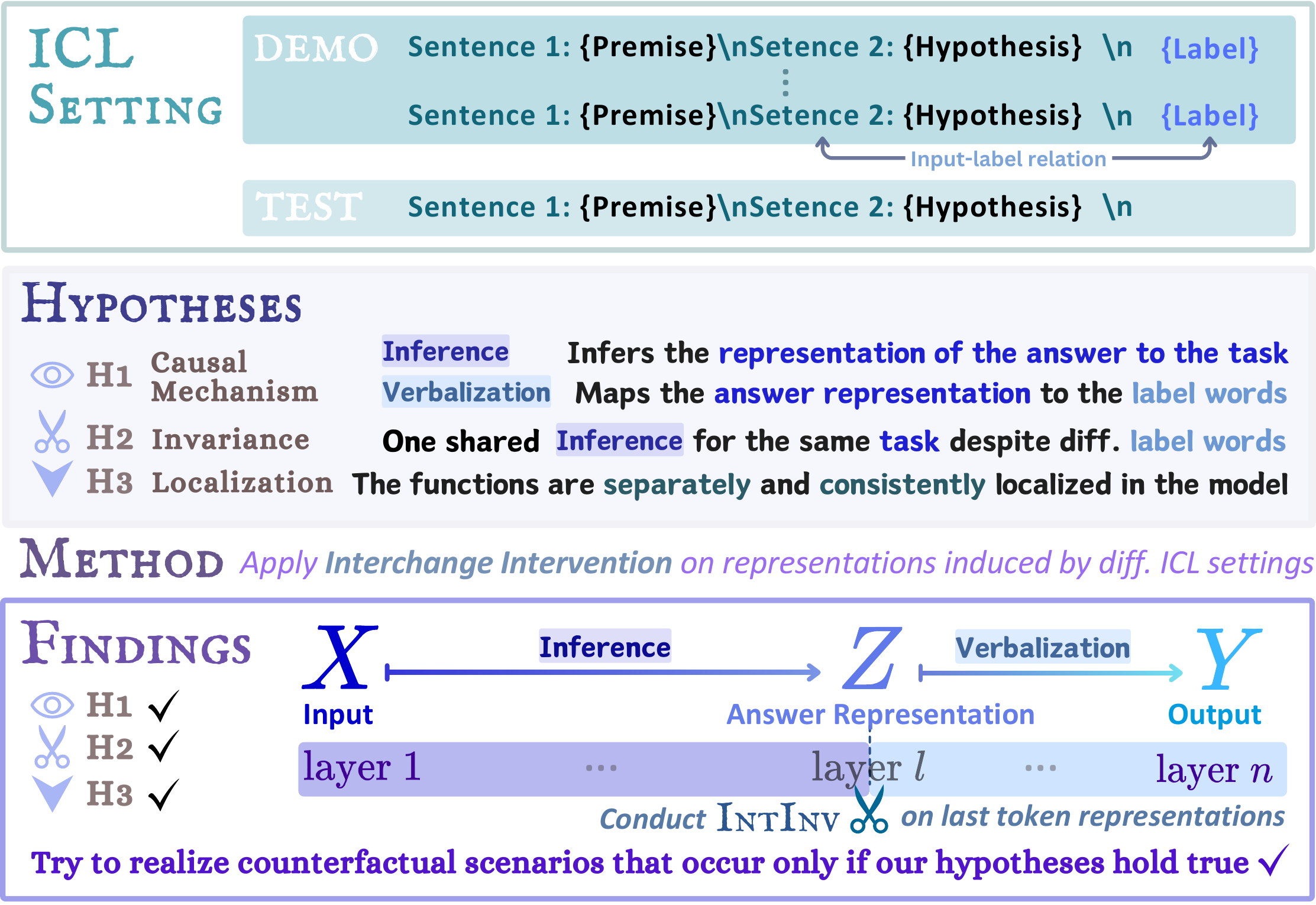}
    \caption{\textbf{Summary of our hypotheses, method, and findings.} We hypothesize that LMs perform ICL via two sequential functions: (1) an \funcinfer that uses an input ICL context and returns a representation of the answer and 2) a \funcverb that maps the aforementioned answer representation to the output label space specified by the demonstrations. Furthermore, the \funcinfer is invariant to remappings of the output label space. We empirically test our hypotheses by conducting interchange intervention experiments (i.e., fixing one of the functions while modifying the other).}
    \label{fig:overview}
\end{figure}

Large language models (LMs) are capable of in-context learning (ICL)—the ability to solve novel tasks from solely a handful of demonstration examples provided in-context during inference \citep{brown2020language}. 
Previous work has found that, in certain settings, ICL performance is minimally affected by using demonstrations with irrelevant label words \citep{min-etal-2022-rethinking}. How do LMs manage to perform in-context learning with irrelevant and even misleading label words?
This research seeks to offer a \textit{causal} explanation for such model behaviors, going beyond mere summaries of input-output patterns. 
We hypothesize that the model develops and utilizes certain abstractions to consistently work well on ICL tasks despite remappings of the label spaces. Specifically, we propose and test the following hypotheses:

\paragraph{$H_1$ Causal mechanism:} LMs sequentially apply two functions when performing ICL: First, an \coloredinfer that constructs a representation of the answer to the ICL input; and second, a \coloredverb that maps this answer representation to the output label space specified by the demonstrations.




\paragraph{\textbf{$H_{2}$ Invariant to label remapping:}} The \coloredinfer is invariant to remappings of the output label space (e.g., ``true''/``false'' to ``cat''/``dog'').



To learn about the model's \textit{low-level} implementations of these \textit{high-level} abstractions (distributed across neural activations), we train probes to detect representations of label words. Its results suggest that the two functions are \textit{separately} located in the \textit{similar} layers across different runs (Appendix~\ref{sec:appendix-motivating-probing-study}). This motivates an additional hypothesis:


\paragraph{$H_{3}$ Separate and consistent localization:} The two functions are located (largely) \textit{separately} in different sets of sequential layers, and the locations of these layers are \textit{consistent} across settings with different remapped label spaces.

This is a working hypothesis that serves as a ``ladder'', providing protocols for us to intervene on the high-level abstractions through modifying the low-level internal representations of the model.

\paragraph{Methods.} 
Our hypotheses are validated if we can define and realize a counterfactual scenario that occurs \textit{only if} our hypotheses are true. For example, let's imagine a scenario where we ``concatenate'' an \funcinfer and a \funcverb, with one induced from a run where the label words are ``true''/``false'', and another from a run where the label words are ``cat''/``dog''. Suppose the output of the former run is ``true'', and the latter run is ``false''.
 This concatenation can make the model generate the hypothetical \coloredcf ``cat'' \textit{only if} all of our hypotheses hold true (see more discussion on this in Section~\ref{operationalize-hyp-test}).
We realize this hypothetical concatenation by swapping the last token representations created for one input into the model which is processing another input with remapped label words.
This operation is known as interchange intervention or activation patching \citep{geiger_nliembed_2020, vig_2020, finlayson-etal-2021-causal, meng_2022}. 

\begin{figure}[t!]
    \centering
    \includegraphics[width=1.0\linewidth]{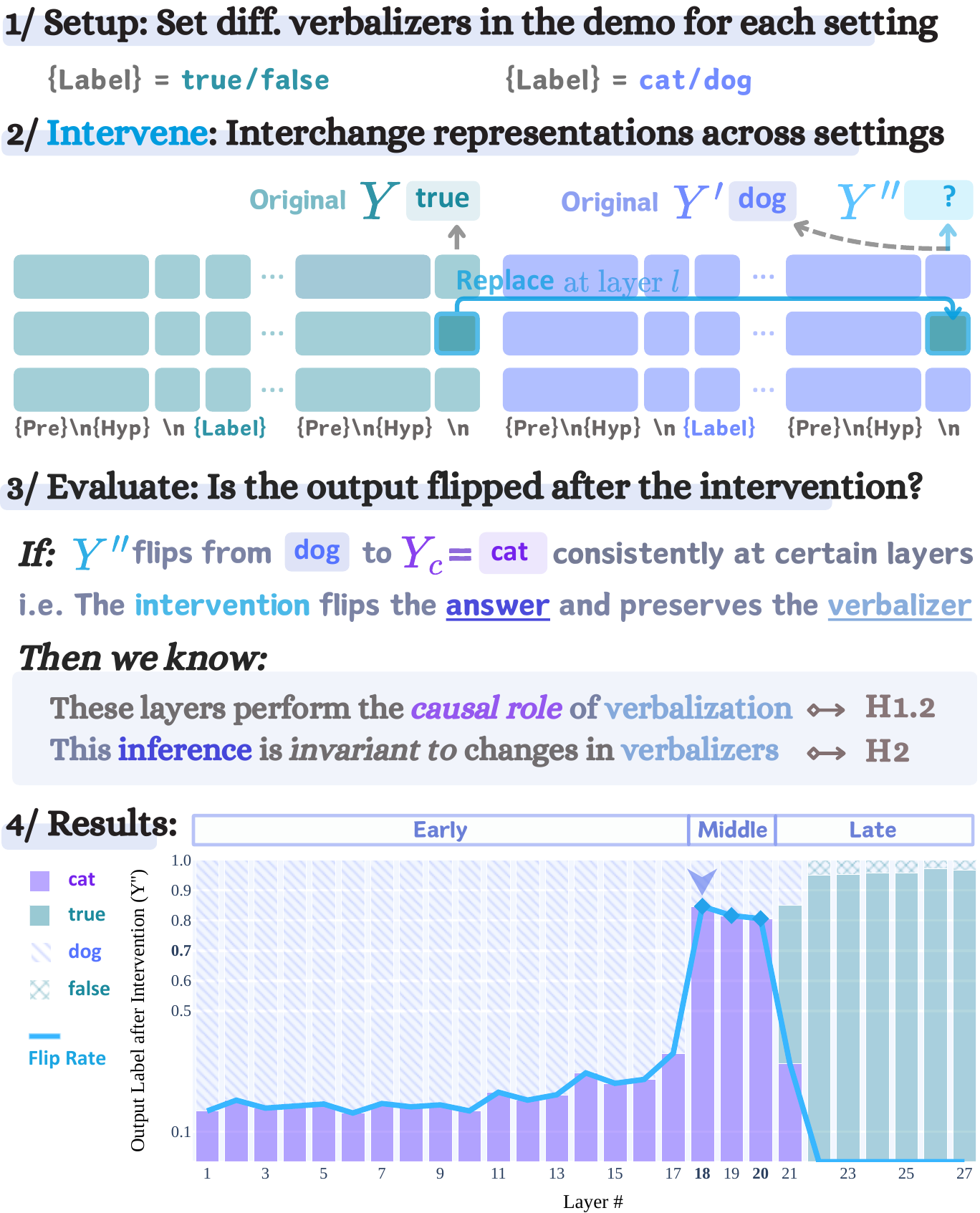}
    \caption{\textbf{Intervention experiments with remapped label spaces.} (1, top): First, we induce changes in the \funcverb by remapping the output label space in the demonstrations (e.g., ``true''/``false'' to ``cat''/``dog'').
    (2): Then, we intervene on the \funcverb by replacing the representation of the {\color{verb-intervened}{intervened model}} (prompted with the remapped label space ``cat''/``dog'') at layer $l$ with the representation from the {\color{verb-source}{source model}} (prompted with the default label space ``true''/``false'') at the same layer.
    (3): We evaluate the intervention effects by measuring how often the intervened output $Y''$ matches the hypothetical \cf $Y_c$.
    (4, bottom): Distribution of outputs predicted by \textsc{Gemma-7b} on MultiNLI when intervening with the remapped label space ``true''/``false'' $\rightarrow$ ``cat''/``dog''. For clarity, we visualize the subset examples where the hypothetical \cf is ``cat''. We observe that the rate at which the intervened output matches the \cf peaks around layers 18-20, localizing the \funcverb and suggesting the \funcinfer is invariant to label remapping.}
    \label{fig:illu-change-verbalizer} 
\end{figure}

\paragraph{Findings.} By experimenting with remapped label spaces that will induce changes in the \coloredverb (illustrated in Figure~\ref{fig:illu-change-verbalizer}), we find that our intervention on certain layers \textit{can} achieve a ``concatenated' model that produces the \cf. This validates $H_2$. We also conduct a complementary experiment to induce changes in the \coloredinfer with constructed alternative tasks on MultiNLI, where the example-label relations are changed and the input and label spaces are fixed (illustrated in Figure~\ref{fig:illu-change-task}). This is aimed for further validating the localization of the \coloredinfer. The results from these two experiments together validate $H_1$. Additionally, we observe that the two functions are \textit{separately} and \textit{consistently} located in similar layers of each model, across various tasks and datasets, including MultiNLI, RTE, ANLI, IMDb, and AGNews. This is aligned with $H_3$. These findings apply to models of various sizes and families, including \textsc{Gemma-7B}, \textsc{Mistral-7B-v0.3}, \textsc{Gemma-2-27B}, and \textsc{Llama-3.1-70B}.\footnote{Our code is publicly available at \url{https://github.com/JunyiTao/infer-then-verbalize-during-icl}.}

\section{Methods}

We aim for an explanation that reveals the causal structure employed by the model. Successful causal explanation is marked by its ability to allow us to make counterfactual predictions, that is, answering the ``what-if-things-had-been-different questions'' should we perform certain interventions \citep{Woodward2003-WOOMTH}. Typically, direct observation of outcomes from distinct interventions on the same unit is not possible  \citep{holland-1986}. But, it is not an issue for us, because we can let the model generate outputs under any conceivable intervention on the units. Moreover, we can make counterfactual statements with mere interventional information with deterministic neural network models \citep{pearl_hierarchy_2022}, which is also typically not possible \citep{pearl_bookwhy_2018}. These enable a concrete operationalization of our hypothesis testing. 

\begin{figure*}
    \centering
    \includegraphics[width=1\textwidth]{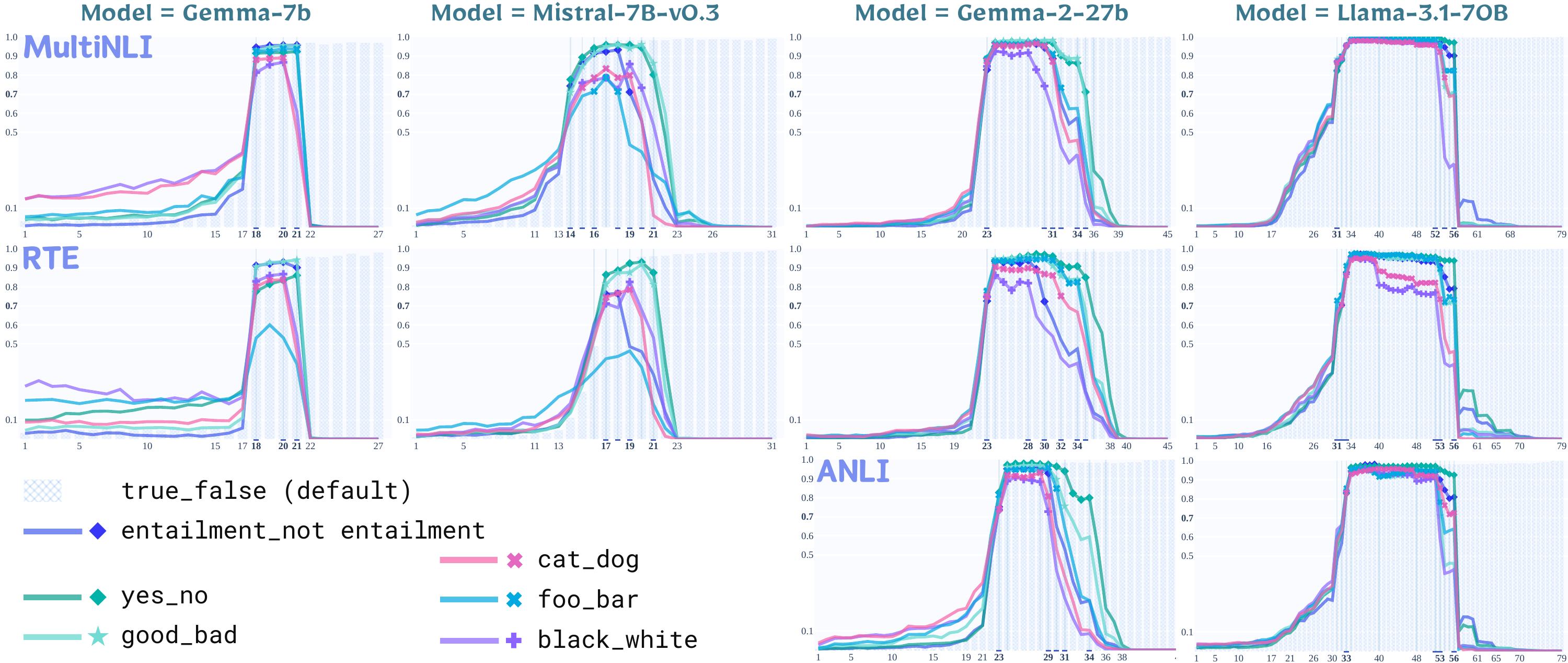}
    \caption{\textbf{NLI experiments with remapped label spaces.} The y-axis is the rate of the output label flipped into the hypothetical \cf label (``flip rate'') and the x-axis is the intervened layer. Each curve represents the flip rates observed when the intervened model is prompted with the corresponding label space. The background bars represent the flip rates of the \textit{default $\gets$ default baseline}, where the intervened model is also prompted with the default label space (``true''/``false''). Due to their unsatisfactory ICL performance, we do not conduct intervention experiments with the two 7-billion models on ANLI (see more discussion on this in Section~\ref{realize-scenario}).}
    \label{fig:results-change-label-nli}
\end{figure*}

\subsection{Testing Hypotheses with Counterfactual Scenarios}
\label{operationalize-hyp-test}

\paragraph{Framework for testing our hypotheses.} Our approach is to (1) define a counterfactual scenario aligned with our hypotheses, wherein certain model behaviors emerge \textit{only if} our hypotheses hold true, (2) employ appropriate intervention methods to realize this scenario, and (3) conduct the intervention-based experiments and interpret their results.

\paragraph{Design a counterfactual scenario.}
\label{design-cf-scenario}


Imagine a scenario corroborating our hypotheses, where we ``concatenate'' the \funcverb and \funcinfer induced from different model runs with distinct label words. Assuming \textbf{$H_2$ (invariant to label remapping)} is true, this concatenation should function effectively, enabling the \funcverb to successfully decode the answer representation produced by the \funcinfer and correctly verbalize it, leading to a hypothetical \cf where the answer from the first run appears in the label words of the second run. Their functionalities must remain unaffected by the concatenation for \funcinfer to transfer the answer representation to \funcverb, collectively producing the counterfactual output.



\paragraph{Operationalize the setting.}

With our ``ladder'' hypotheses, we can modify the model's \textit{low-level} internal representation as a protocol for intervening on the \textit{high-level }\funcinfer and \funcverb. Specifically, the two functions being localized \textit{separately} (\textbf{$H_{3.1}$ separate localization}) enables us to isolate their causal effects, and \textbf{$H_{3.2}$ (consistent localization)} makes it possible to conduct the intervention with the counterfactual value of the representation—the value it would take under alternative input scenarios \textit{in the same position}—rather than a value the neuron might never \textit{actually} assume. With these, our objective becomes to find a layer (or sequence of layers),  $l_{mid}$, such that taking the last token representation from one run and plugging it into another would change the model’s output in a way that reflects the hypothetical counterfactual condition.

\paragraph{Interpreting results.} Our hypotheses are supported if the model consistently generates a sufficient proportion of \texttt{counterfactual outputs}. Additionally, this will also provide information about the specific (though not exact) locations of the two functions within the model—$l_{mid}$ should occur after the layer where the answer representation is sufficiently developed. This, however, does not tell us the exact starts and ends of the two functions. It is possible that only a \textit{subset} of layers before $l_{mid}$ is actively involved in implementing the \funcinfer and after $l_{mid}$ is actively involved in implementing the \funcverb, while some others may be redundant. 

Our hypotheses can be weakened by the absence of evidence, i.e., no specific layer found to allow effective concatenation of the functions; and they can falsified if our interventions yield consistent results that contradict our predictions, which may indicate that the model systematically employs mechanisms different than those hypothesized. 




\subsection{Using Interchange Interventions to Realize Counterfactual Scenarios}
\label{realize-scenario}

\paragraph{Interchange intervention.} We will first introduce the formulation of interchange intervention \citep{geiger_nliembed_2020, geiger_causalabstraction_2021, geiger_inducingcausalstructure_2022, geiger_causal_abstraction_2024,geiger_findalignment_2024, huang_ravel_2024} and then apply it to our context.
Consider a model $\model$ that takes an input string $\mathbf{x}$ and generates an output string $\mathbf{y}$. We denote the entire set of internal representations of the model $\model$ created during this inference as $\model(\inputx)$ and the predicted token  $\mathbf{y} = \predfunc(\model(\inputx))$. 

Let's conduct interchange intervention on a set of intermediate representations $Z$ of the model $\model$ by replacing them with the value of $\mathbf{z}$ and denote the post-intervention model as  $\model_{\mathbf{Z} \gets \mathbf{z}}$. The difference between $\predfunc(\model(\inputx))$ and $\predfunc(\model_{\mathbf{Z} \gets \mathbf{z}}(\inputx))$ manifests the causal contribution of $\mathbf{Z}$ to the model behavior. To get the intended value of $\mathbf{z}$, we use \texttt{GetVals} to retrieve the representation values that the variables $\mathbf{Z}$ \textit{would have} taken on if $\model$ processes another input $\inputx'$, denoted as $\getvals{\mathcal{M}}{\inputx'}{\mathbf{Z}}$. With the two operations, we obtain the post-intervention version of the model, \({{\mathcal{M}}_{\mathbf{Z} \leftarrow \getvals{\mathcal{M}}{\inputx'}{\mathbf{Z}}}}\), where the values of $\mathbf{Z}$ set as those obtained by processing $\inputx'$. We refer to  $\model(\inputx')$ as the \textbf{source model} since it provides the desired values of $\mathbf{z}$ and $\model(\inputx)$ as the \textbf{intervened model} whose internal representations are to be modified. Thus, we get the output of the intervened model after the interchange intervention: \(\label{eq:intinv}
\intinv(\mathcal{M}, \inputx, \inputx', \mathbf{Z})
\coloneq 
\predfunc\big({{\mathcal{M}}_{\mathbf{Z} \leftarrow \getvals{\mathcal{M}}{\inputx'}{\mathbf{Z}}}}(\inputx)\big)
\).

In our study, we want the intervention to reroute the computational graph of the source model to the intervened model during the last $l$ layers, to achieve a ``concatenation'' of functions from different runs (illustrated in Figure~\ref{fig:illu-change-verbalizer}). We achieve this by setting the variables $\mathbf{Z}$ to be the representations of the \textit{last token} of the input context. The causal attention mask of the decoder-only architecture of the LMs we study ensures that no intermediate representation of a token is influenced by its successors, allowing interventions on the last token without altering the computational graph of preceding tokens. Thus, the intervened model will apply the same function during the last $l$ layers for both its original representation and the plugged-in representation.

\paragraph{Formulate counterfactual outputs.}

An ICL prompt $\mathbf{x}$ is composed of (1) a set of demonstrations $\{x_{demo}^{(j)}, y_{demo}^{(j)} \}_{j=1}^{k}$, where the example-label relations are intended to determine the task to be inferred, such as NLI, and $y_{demo}^{(j)}$ in the output label space $\mathcal{Y}$, such as $\{\texttt{"true"}\text{,}\texttt{"false"}\}$ where each element corresponds to the positive and negative label respectively, (2) a test example $x_{test}$, and (3) a prompt template (see Figure~\ref{fig:overview}, Top). 

In Section~\ref{design-cf-scenario}, we imagine a counterfactual scenario where we concatenate two functions, each induced by $\model(\inputx)$ and $\model(\inputx')$. The hypothetical \cf here reflects that the answer comes from the model $\model$ inferring $\inputx'$ and appears in the label words specified in $\inputx$. Note that when experimenting with \textbf{remapped label spaces} (Figure~\ref{fig:illu-change-verbalizer} and Section~\ref{change-verbalizer}), we take the \textit{upstream} \funcinfer from the source model (with default label words) to pass information to the \textit{downstream} \funcverb of the intervened model, \textit{not the reverse}.


To make the causal effects of our intervention \textit{observable} on the behavioral level, we set up $(x,x')$ pairs to ensure that the hypothetical \cf always \textit{differs} from the original output of the intervened model. For example (illustrated in Figure~\ref{fig:illu-change-verbalizer}), consider an input $\inputx'$ with the default label words ``true''/``false'' and an input $\inputx$ with the irrelevant label words ``cat''/``dog''. If feeding $\inputx'$ to the source model yields an output $\predfunc(\model(\inputx')) = \texttt{true}$, we will sample the corresponding $\inputx$ to construct our intervention experiment from the subset $\{\mathbf{x} \mid \predfunc(\model(\inputx)) = \texttt{dog} \}$ where the intervened model outputs ``dog''. This enables us to observe whether the output of the intervened model \textit{changes} from ``dog'' to the \cf ``cat''. This construction requires the test example in $\mathbf{x}$ and $\mathbf{x}'$ to differ from each other, which we will justify later in this section. 

One interesting property from the ``concatenation'' view is that we can interpret the ``flip'' from two perspectives: (1) the output of the intervened model flips from ``dog'' to ``cat'', manifesting the causal effects of \funcinfer, or (2) the output of the source model flips from ``true'' to ``cat'', manifesting the causal effects of \funcverb. 


\paragraph{Evaluate causal effects.}
We measure the effectiveness of our interchange intervention based on the \textbf{flip rate}, which is straightforwardly defined as the percentage of cases in which the intervention succeeds in making the intervened model's output flipped to the hypothetical \cf. The strength of the evidence supporting our hypotheses is proportional to the flip rate: $\geq 70\%$ is considered adequate, and $\geq 90\%$ is considered very strong.

We further construct a \textit{\textbf{default $\gets$ default baseline}} where the intervention is conducted on the intervened model and the source model that are both prompted with the default label words. The \cf is simply a flipped label word: if the intervened model's original output is ``true'', then the \cf is ``false''. 
We consider the processes the model implements at certain layers to be \textit{shared} across settings with remapped label spaces if the corresponding flip rates are \textit{close} to this baseline; conversely, if the flip rates \textit{diverge} from this baseline, we conclude that the model uses different functions during these layers. In this case, the diverging processes are very likely to perform causal roles related to the label words. 

\paragraph{Use different test examples.} Another implication of differentiating the test examples in $\inputx$ and $\inputx'$ is it prevents the intervened model from ignoring the plugged-in representations and just solving the task on its own by attending to the test example in its input contexts. We control that only the source model ``knows'' the test example, ensuring that the concatenated model produces the \cf \textit{only if} it correctly verbalizes the answer passed on by the \funcinfer. 

\paragraph{Filtering out cases with low ICL accuracy.}
We generally focus on a subset of cases where the model achieves adequate ICL accuracy (above $0.7$). This serves dual purposes: first, to verify if these results are reproducible on \textit{base} models, and second, to ensure that the model’s successes are not merely due to chance, but rather a result of systematic causal mechanisms developed to address the task. We discuss how we adjust the prompting strategy to improve the ICL performance without explicit instructions in Appendix~\ref{prompting-strategy} and explore the low ICL cases in Appendix~\ref{interpret-negative-cases}.

\section{Experiments}

\paragraph{Models.} We choose open-sourced base models of various sizes and families, including \textsc{Gemma-7B} \citep{team2024gemma}, \textsc{Mistral-7B-v0.3} \citep{jiang2023mistral}, \textsc{Gemma-2-27B} \citep{team2024gemma}, and \textsc{Llama-3.1-70B} \citep{dubey2024llama}. 

\paragraph{Datasets.} We first conduct experiments on three NLI datasets: MultiNLI \citep{williams2017broadcoverage}, RTE \citep{wang2018glue}, and ANLI \citep{nie2019adversarial}. 
We cast MultiNLI and ANLI into binary classification tasks by dropping the ``neutral'' class to facilitate the remapping of label spaces and to avoid ambiguities inherent in the ``neutral'' examples. We further test if our findings generalize to other tasks by using two sentence classification datasets: IMDb \citep{maas-EtAl:2011:ACL-HLT2011} and AGNews \citep{Zhang2015CharacterlevelCN}. Due to computational limits, we only take a subset of 300 test examples that are sampled randomly and fixed across different settings on the dataset (see Table~\ref{tab:dataset_stat} for details). 

\paragraph{Implementation details.} We conduct intervention on each layer in the model except the last one, since exchanging the last token representation at that layer is equivalent to replacing the output token. ICL settings are summarized in Table~\ref{tab:icL_setting}. All flip rates are averaged over three runs with evenly sampled demonstration examples, i.e., balanced for each class. Model outputs are decoded greedily by always selecting the top predicted token.

\begin{figure*}[t]
    \centering
    \includegraphics[width=1\textwidth]{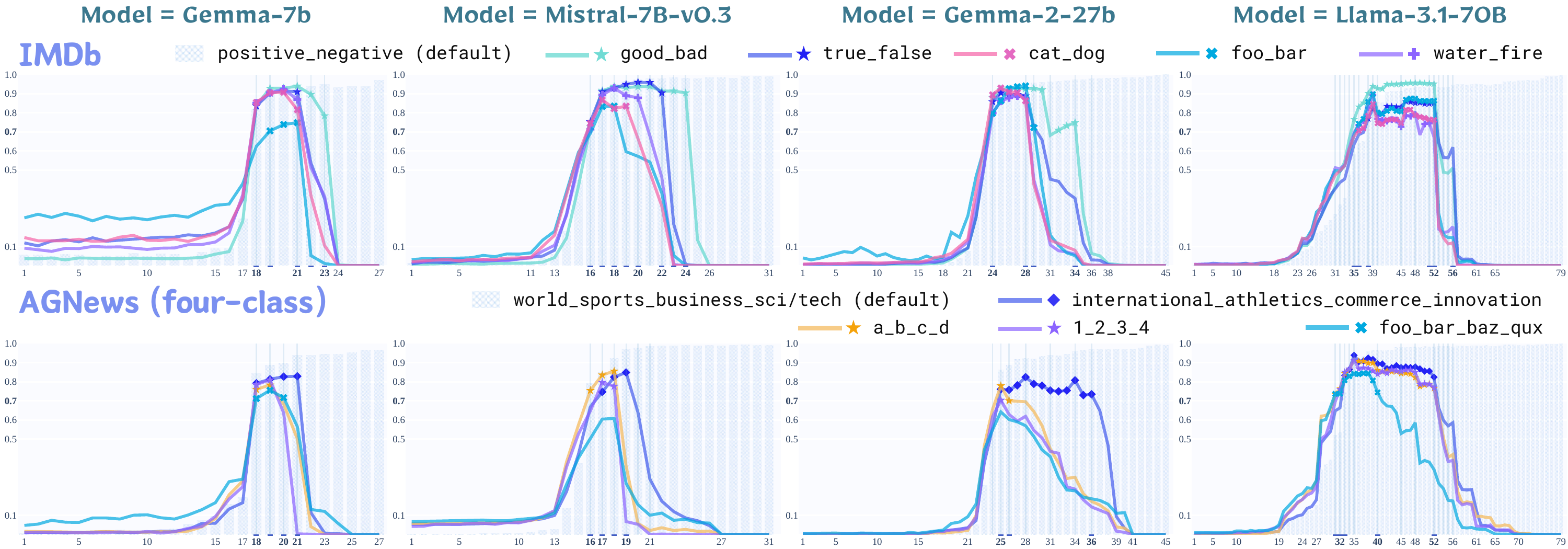}
    \caption{\textbf{Experiment with remapped label spaces on IMDb and AGNews.} We perform the same experiments with remapped label spaces on tasks other than NLI. Different sets of label spaces are used for IMDb and AGNews to accommodate the semantic variations of the label words.}
    \label{fig:results-change-label-imdb-agnews}
\end{figure*}

\subsection{Experiment with Remapped Label Spaces}
\label{change-verbalizer}

\paragraph{Modulate the label words.} For each task, we experiment with a diverse set of label words. For example, in addition to the default label words used for NLI (``true''/``false''), we select (1) the task-related label words ``entailment''/``not entailment'' and (2) a set of task-irrelevant label words, including those generally conceived and used as binary pairs (``yes''/``no'' and ``good''/``bad'') and those that are less obviously binary or used in contrasting contexts (``cat''/``dog'', ``foo''/``bar'', and ``black''/``white''). We list all label words in Table~\ref{tab:results-single-tokens}.

\paragraph{Findings of counterfactual outputs.} 
We will examine both the patterns of flip rates (Figure~\ref{fig:results-change-label-nli}) and the distribution of output tokens (Figure~\ref{fig:illu-change-verbalizer}, bottom; Appendix~\ref{sec:appendix-output-token}). Below, we discuss these results based on a representative setting: the experiments on \textsc{Gemma-7b}, as illustrated in Figure~\ref{fig:illu-change-verbalizer}. 

During the early layers (1-17), flip rates are generally very low.  Such ineffective intervention suggests that the answer representation is not yet sufficiently developed to be decoded by the other \funcverb. This can be validated by the distribution of output tokens: the intervened model mostly generates ``dog'', which is the original output of the intervened model.

During the middle layers (18-20), we observe that nearly all settings achieve a flip rate of at least $70\%$. This means our interventions on these layers can effectively achieve the counterfactual scenario, that is, create a ``concatenated'' model that outputs the \cf ``cat''. This success depends on the \funcverb of the intervened model correctly decoding representations from the \funcinfer of the source model and maintaining a correct mapping function to the label space.

During the late layers (21-27), the model outputs are gradually dominated by the original outputs of the source model (``true'' instead of ``cat''). It is possible that the intervention may have bypassed the starting point of the \funcverb, which means the representations can no longer be successfully transformed into the intended label space. 


\paragraph{Findings of generalizability.} We experiment with three groups of tasks and datasets, including: the primary setting MultiNLI (Figure~\ref{fig:results-change-label-nli}, Top), other NLI tasks such as RTE and ANLI (Figure~\ref{fig:results-change-label-nli}, Mid and bottom), and other tasks such as sentiment analysis on IMDb and topic classification on AGNews (Figure~\ref{fig:results-change-label-imdb-agnews}). 
The patterns' generalization to other NLI datasets means they are not just specific to some constructs in MultiNLI; the generalization to IMDb means that they are not specific to NLI but perhaps to classification tasks in general; and the generalization to AGNews means that they are not specific to binary classification tasks but are shared even in multi-class classifications.

\begin{figure*}
    \centering
    \includegraphics[width=1\textwidth]{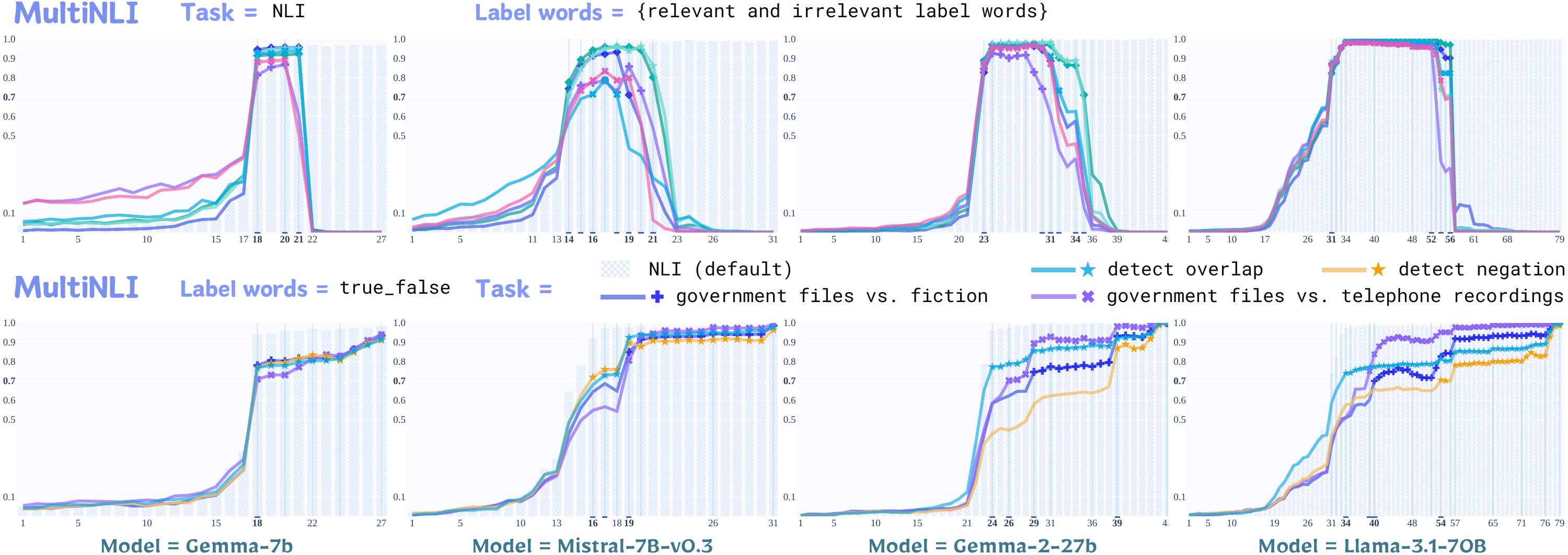}
    \caption{\textbf{Experiment with reconstructed tasks on MultiNLI.} Curves represent the flip rates of different settings where the intervened model is prompted with alternative tasks. The intervened model always sees the default task. The background bars represent the flip rates of \textit{default $\gets$ default baseline}, where both the intervened model and the base model are prompted with the default task (NLI). }
    \label{fig:results-change-task}
\end{figure*}

\paragraph{Caveat.} An alternative explanation for the results is that the intervened model ignores the plugged-in representations entirely and just re-solves the task on its own, given that the plugged-in representation may fully represent the test example. In this case, what has been transferred in the intervened representations is merely a task-agnostic representation of the test example. To address this concern, we conduct a complementary experiment to induce changes in the \funcinfer with constructed alternative tasks on MultiNLI, where the example-label relations are changed and the input and label spaces are fixed (see Section~\ref{sec:change-task}). So far we can only conclude that the computation after the middle layers, e.g., after layer 18 for \textsc{Gemma-7b}, at least performs \texttt{verbalization}. 

\paragraph{Further interpretation of the results of multi-class classification.} Flip rates sometimes decrease \textit{earlier} on AGNews than in binary classification. Why? We analyze the distribution of output tokens (Appendix~\ref{sec:appendix-output-token}). In binary classification datasets, the output tokens are typically dominated by the \cf during middle layers, which we labeled as ``flipped'' tokens; subsequently, these tokens are increasingly replaced by those from the source model, which we refer to as ``overwritten'' tokens. However, on AGNews, we observe that during the middle layers: (1) a significant proportion of ``flipped'' tokens become ``overwritten'' tokens, and (2) the remainder are tokens not predefined as default or irrelevant, referred to here as ``other tokens'' (Figure~\ref{fig:token-agnews1} and Figure~\ref{fig:token-agnews2}).

This behavior does not contradict our hypotheses. It is possible that the \funcinfer is implemented in the same layer as in binary settings, evidenced by the peak in flip rates around similar layers in both binary and AGNews classification; while the \funcverb starts in earlier layers in multi-class classification, whose initialization is marked by the increasing dominance of ``overwritten'' tokens. Meanwhile, a further investigation into the ``other tokens'' reveals that they seem to still encode the correct information of the answer and be decoded into tokens that are semantically similar to the default ones. For example, ``other tokens'' includes  ``science'' and ``technology'', both are synonymous with the default label word ``sci/tech'', and the token ``politics'' will pick up examples similar to those labeled by ``world'' in the default label space. Together, the findings suggest that the intervened model can still \textit{correctly decode} the answer representation transmitted by the \funcinfer of the source model, but its implementation of the \funcverb is more sensitive to our interventions. 

\subsection{Experiment with Reconstructed Tasks on MultiNLI}
\label{sec:change-task}
\paragraph{Purpose.}
To address the potential concern mentioned earlier, we design a complementary experiment to test if the plugged-in representations are task-\textit{agnostic}. We want to control that the intervened model can \textit{only} know about the intended tasks (that can lead to the \cf) by decoding a task-\textit{relevant} representation from the source model, i.e., the representations we take from the source model and plug into the intervened model. We conclude that the plugged-in representations are \textit{not task-agnostic} if the model can still achieve high flip rates; and this conclusion can be strengthened if the high flip rates occur in a similar location as in the previous experiments with remapped label spaces, that is, the results validate the location of the \funcinfer.

\paragraph{Set desiderata for alternative tasks.}
We want the alternative tasks that (1) maximally preserve the input space to allow the model to continue employing similar mechanisms, (2) secure a balanced dataset with an adequate number of examples for both positive and negative labels, and (3) allow the model to attain sufficient ICL accuracy, which means the model has been redirected to the new tasks and developed corresponding abstractions. The criterion for ICL accuracy is less stringent than the experiment with remapped label spaces given the challenges of re-purposing a dataset designed for another task.


\paragraph{Construct alternative tasks.}
To achieve these conditions, we want to find suitable tasks that are typically simple, deterministic functions, especially those based on shallow linguistic features or those that capture the ambiguity or heuristics exploited by the model. Following \citet{si-etal-2023-measuring}, beyond the default task NLI, we construct two domain classification tasks using existing metadata of the MultiNLI and two lexical classification tasks reflecting the simple syntactic heuristics the model may exploit to achieve high performance on NLI tasks \citep{mccoy-etal-2019-right}. We list them in Table~\ref{tab:ambi-task}.

\paragraph{Apply interchange Intervention.} In experiments, we always feed the NLI task to the intervened model and use ``true''/``false'' for the label space (illustrated in Figure~\ref{fig:illu-change-task}). We define different tasks with the same sets of demonstration examples, but use different example-label relations for each task. Specifically, each data point is assigned with two labels, with one for the NLI task and another for the alternative task. In addition, we prepend instructions to each example, adopted from \citep{si-etal-2023-measuring} (summarized in Table~\ref{tab:ambi-instruction}), since none of the prompting templates without explicit instructions we have tried can lead to adequate ICL performance. The \textit{default $\gets$ default baseline} is created by using the NLI task for both the intervened and the source model.

\paragraph{Findings of consistent locations.} As shown in Figure~\ref{fig:results-change-task}, the flip rates surge at the same layer in both the experiment with remapped label spaces and with reconstructed tasks. Given that the intervention design ensures only the source model has access to the correct task—which leads to the hypothetical \texttt{counterfactual output}—a high flip rate is possible only if the plugged-in representation has already encoded information about this \cf. This means that the early-stage layers of the upstream source model have already performed the task, i.e., implemented the \funcinfer. Thus, the results validate the location of the \funcinfer and thus help us address the aforementioned concern.

\paragraph{Caveat.} The evidence coming from this experiment will only serve complementary roles, and its strength is weaker than the experiment with remapped label words. First, it is hard to control what has been changed when intervening on the upstream process; and second, the patterns here are less consistent (alternative tasks can have flip rates varying in values and locations where they peak). However, these do not contradict our hypotheses, as they do not require high and consistent flip rates in this setting. The goal of this experiment is to verify the early-stage layers do perform \funcinfer, and for this purpose, it is sufficient to observe that in all settings (except one for \textsc{Mistral-7B-v0.3}), the flip rates with reconstructed tasks rise significantly above the random baseline ($0.5$) before the flip rates in the experiment with remapped label spaces drop to $0$. The inconsistencies of the locations where the flip rates peak are expected, since changing the task likely causes significant changes in the model's internal circuits and states used to solve the tasks (even though these tasks might share some sub-processes), making the representations less transferable. 


\section{Discussion}

\paragraph{Validation of $H_1$: Causal mechanism.}  
This hypothesis is confirmed if we can ``concatenate'' an \funcinfer with another \funcverb and observe the generation of the \cf that reflects the answer obtained by the \funcinfer and the corresponding label space of the \funcverb. We achieve such a scenario with the experiment with remapped label spaces, as manifested by the high flip rates we observe during the middle layers of the model. 

Concern may be raised that the intervened model just re-solves the task on its own during the downstream processes, that is, both \funcinfer and \funcverb are performed in the middle and the late layers. Our experiment with alternative tasks shows that this is not true. This experiment ensures that only the source model knows the answer that leads to the hypothetical \cf, which means high flip rates can only be achieved if the plugged-in representation involves a sufficiently developed representation of the answer. Therefore, the high flip rates we observe during and after the middle layers support the hypothesis that the \funcinfer is implemented by the upstream processes (before the layers with effective interventions).

\paragraph{Validation of $H_2$: Invariant to label remapping.} From the results of the experiment with remapped label spaces, we know that intervening on a sequence of middle layers in the model can successfully achieve the counterfactual scenario, where the \funcinfer can be concatenated with another \funcverb to produce the hypothetical \cf. This indicates that the \funcinfer can play its causal role invariant to the downstream \funcverb.

\paragraph{Validation of $H_{3}$: Separate localization and consistent localization.} This hypothesis is a necessary condition for us to use interchange intervention to realize the counterfactual scenario, and thus it gets validated if our interventions are effective. It can be further supported by the observation that for each model, the trend of flip rates is similar across different datasets and tasks. For example, flip rates surge at the same or very close layers (Figure~\ref{fig:results-change-label-nli} and~\ref{fig:results-change-label-imdb-agnews}). We summarize the start and the end of the layers enabling effective interventions in Table \ref{tab:middle_stage}.


\section{Related Work}



\paragraph{Understanding ICL.}
A variety of work has sought to better understand how language models perform in-context learning. \citet{min-etal-2022-rethinking} show that ICL performance on a variety of text classification tasks is minimally affected by randomly changing the demonstration labels, hypothesizing that demonstrations may primarily help with determining the label space (i.e., the \funcverb), the distribution of the input examples, and how the output text should be formatted.
\citep{DBLP:conf/iclr/XieRL022} hypothesize that LMs learn latent tasks during pre-training, primarily using ICL demonstrations to identify which latent task is most pertinent to the provided ICL input.
Other work \citep{akyrek2022learning,oswald2022transformers,dai2022gpt} hypothesizes that LMs perform implicit gradient descent on a latent model during ICL.
\citet{wei2023larger} and \citet{DBLP:conf/acl/PanG0C23} confirmed these observations and further indicated that this consistency depends on the size of the model. Our study offers a deeper analysis of these behaviors at the representational level.
\citet{wei2023larger} study how model size affects whether LMs prefer to rely on semantic priors about a task, as opposed to the example-label mappings provided in the demonstrations. They find that smaller models tend to rely on semantic priors during pre-training, since they ignore flipped labels presented in the ICL demonstrations. In contrast, larger models make use these flipped labels, indicating that they rely less on their semantic task priors.
\citet{DBLP:conf/acl/PanG0C23} disentangle task learning and task recognition in in-context learning; intuitively, LMs use their priors to perform novel tasks during ICL (since a few demonstrations are unlikely to provide complete information about a complex task), but they also do not completely rely on their task priors (since they can handle ICL settings with arbitrary label mappings).

\paragraph{Implicit functions in language models.} Another line of prior work has sought to characterize what functions might be implicitly implemented by LMs. \citet{olsson2022incontext} identify ``induction heads'', a unique type of attention head that replicates repeating patterns from prior contexts. \citet{hendel2023incontext} argue ICL compresses the demonstrations into a task vector, which is then used to generate the output for the ICL input. \citet{merullo2023mechanism} provide evidence that Transformer LMs perform ICL in three layer-wise stages: argument formulation, function application, and saturation.
\citet{todd2024functionvectorslargelanguage} show that ICL creates function vectors of the demonstrated task, and these function vectors can trigger execution of the task in other settings (e.g., zero-shot prediction or natural text).

\section{Conclusion}

We hypothesize that when LMs perform ICL with irrelevant or misleading label words, they first apply an \coloredinfer to obtain a representation of the answer and then apply a \coloredverb to verbalize the answer as one of the label words specified in the demonstrations ($H_1$). In addition, the \coloredinfer is invariant to remappings of label words ($H_2$), and both the functions can be localized separately and consistently within certain layers of the model ($H_3$). To validate our hypotheses, we design experiments based on interchange intervention to realize counterfactual scenarios that would \textit{only} occur if all of our hypotheses hold true. Our experiments across a variety of tasks, datasets and models indicate that our hypotheses hold across a variety of settings. Our findings contribute to a growing body of work on mechanistically understanding how language models perform in-context learning.


\section{Limitations}

\paragraph{Coverage of models and tasks.} Since our intervention study requires access to the model's intermediate representations, we can only experiment with open-source models. There is no guarantee that similar findings can be observed in the state-of-the-art close models, and how the MoE architectures will affect our findings remains unclear. In addition, we acknowledged that we only focus on classification tasks on which the notion of ``verbalization'' or remapping of label spaces can be more naturally defined. Future studies could extend to more complex tasks, such as those that involve answering open-ended questions.

\paragraph{Further specifying the inference function.} Does the model \textit{truly} carry out natural language inference when processing datasets like MultiNLI, RTE, and ANLI? Our current evidence does not allow us to definitively answer this question, and the notion of ``textual entailment'' itself is not without its controversies. We cannot claim about what the model is actually doing during the layers of the \funcinfer. Moreover, when experimenting with reconstructed tasks on MultiNLI, we do not know the \textit{true} relationships between the alternative tasks (lexical and domain classification) and NLI. Nonetheless, these ambiguities do not undermine our argument—we do not assert that the model performs an ``NLI-inference'' function on NLI datasets. By \funcinfer we refer to the (general) processes the model uses to derive the answer representation.

\paragraph{Predicting ICL performance.} Prior studies have shown that the model's ICL performance can be (sometimes significantly) harmed by the irrelevant labels \citep{min-etal-2022-rethinking, DBLP:conf/acl/PanG0C23, wei2023larger}. We do not conduct a systematic analysis and explanation of how different factors—such as prompt templates, label word choices, and the number of shots—causally contribute to the model's performance on ICL tasks with irrelevant labels. A nuanced understanding of the underlying mechanism would require analysis tailored to each model and setting. In particular, our current research does not address which specific features of label words influence ICL performance positively or negatively, which could be an interesting topic to explore.

\paragraph{Unknown training data.} Concern arises whether the model's good performance on ICL tasks with irrelevant label words stems from its prior exposure to similar cases. We cannot rule out the possibility of data leakage during pre-training, primarily because we do not have access to the pretraining data of the models we study. We are hopeful, however, that our use of a diverse range of irrelevant labels in our prompts reduces the likelihood that these specific cases were present in the pre-training dataset, thus alleviating potential concerns.

\section{Acknowledgements}
We would like to thank Atticus Geiger, Jing Huang, Jacqueline Harding, Zhengxuan Wu, Kara Schechtman, Junlin Wang, Jiarui Lu, Zhihan Li, and the anonymous reviewers for feedback and discussions that helped improve this work. This research was enabled in part by the computational resources provided by Mila (\href{https://mila.quebec}{https://mila.quebec}) and NVIDIA.

\bibliography{references}


\clearpage

\appendix

\section{Probing Study}
\label{sec:appendix-motivating-probing-study}

\begin{figure}[!htb]
    \centering
    \includegraphics[width=1.0\linewidth]{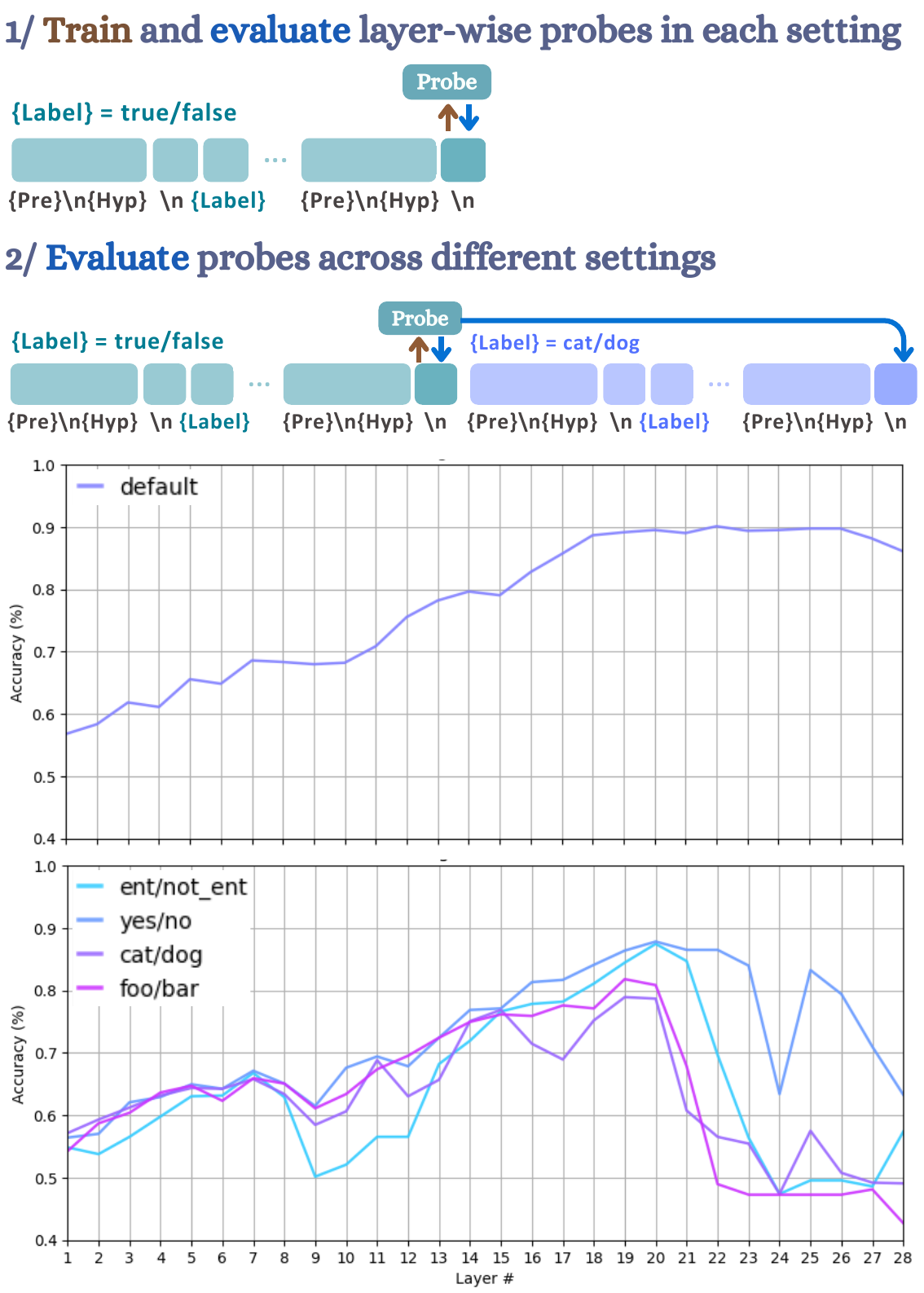}
    \caption{\textbf{Probing setting and results.} \textbf{Top:} an illustration of the probing experiment, where we train a probe on the last token presentation to predict the NLI label. \textbf{Bottom:} results of the probing study. The first graph shows the probing accuracy when the probe is trained and evaluated on the default setting; while the second graph displays the probing accuracy when the probe is trained on the default setting and evaluated on other settings with different label words. By comparing the out-of-distribution curves (in the second graph) with the in-distribution curve (in the first graph), we can see an obvious bifurcation starting around the 18th-20th layer. This indicates (1) that the representation of the answer has been fully developed around these layers, and (2) the first process (before the bifurcation point) is roughly invariant to the changes in label space, while the second process (after the bifurcation point) is heavily dependent on the label space.}
    \label{fig:probing}
\end{figure}
To identify the low-level implementations that align with our hypothesized high-level abstractions, we start with identifying processes \textit{shared} across different model runs and then verify if they satisfy all properties we hypothesize. As a proxy, we conduct a pilot study where we use probing to examine how the intermediate representations, i.e., the neural activations, correlate with the high-level concept of the answer to the ICL task and the label words specified in the prompt. 

\paragraph{Probe.} A probe \citep{alain2016understanding, peters-etal-2018-dissecting,tenney-etal-2019-bert,clark-etal-2019-bert, Hupkes_2018} is usually a linear classifier or shallow MLP that takes the intermediate representation as the input and output labels for some property, aimed at testing how easily the representations can be linearly separated. A high probing accuracy on a hold-out test set indicates that the information about the property is encoded in the intermediate representation.

\paragraph{Experimental details.} We implement probes as logistic regressions with Scikit-learn \citep{scikit-learn} and the L-BFGS optimization algorithm \citep{liu_limited_1989}. We train one probe for each layer with representations generated on the RTE training set by \textsc{Gemma-7b}. Probes are then applied to the RTE validation set with different label words. Results are averaged over three trials with different sets of demonstrations. We repeat this process for each layer.

\paragraph{Experiments.} We train probes to predict the output labels on the last token representations of one run with default label words (``true''/``false'') in the NLI task. We first apply to probe to the representations produced by runs with default label words to observe the development of the representation of the output throughout the forward pass. Then, we test if this probe can generalize to the representations produced by runs with remapped label words (e.g., when the default label space ``true''/``false'' is remapped to ``cat''/``dog''). This aims to identify if there are shared processes in runs with different label words. Specifically, the process on certain layers is interpreted as being \textit{shared} if the probing accuracy on these layers are similar across different runs; on the contrary, if the probing accuracy on certain layers of one run diverges from the probing accuracy on another run, we conclude that the model implements different processes on these layers in each run.

\paragraph{Findings.} For probes trained and tested on the runs with default label words (Figure~\ref{fig:probing}, Middle), the probing accuracy is monotonically increasing, suggesting the development of the representation of the output label.
For probes trained on the runs with default label words and tested on runs with different label words, the probes achieve high accuracy on other label words in layers 15-20 and then drop to a random baseline in the last 8 layers. In other words, the probing accuracy on different runs first converges and then diverges. 

These results indicate that the information about the NLI labels is encoded at least in two different patterns across the model's forward pass. In the middle layer, where the probe can generalize, the representations share a similar structure in relation to the NLI label, enabling the probe trained on the default label words to still correctly classify in the other label words cases. In the late layers, the representation structures start to vary based on the label words, causing a decline in generalization performance. From these results, we can categorize the model's intermediate representations into two groups: one that is sensitive to label words and another that is not. Given these findings, it is natural to assume that the processes that generate the representations share the same characteristics. That is, the first process implemented by the middle layers is \coloredinfer related since it produces unified representations of the task target. The second process implemented by the late layers is \coloredverb related, as its intermediate representation structure varies across different label words. 

\paragraph{Caveat.} We note that probing results only tell us whether the representations encode certain information without guaranteeing that these representations indeed play \textit{causal roles} in the generation of certain model's behaviors, i.e., \textit{used} by the model \citep{belinkov-2022-probing}, or \textit{genuinely represented} in the layers where they are detected \citep{harding_operationalising_2023}. Claiming about the causal effects we want to study will require further hypotheses and experiments introduced later in the paper. Nonetheless, we take the probing results as an initial point, and the three key messages delivered: 
(1) the two functions may be located \textit{separately}, (2) if located separately, the \funcinfer comes first then the \funcverb, and (3) the location of the two functions may be \textit{consistent} across settings.

\paragraph{Clarification of ``localization''.}
We do not commit to the notion of ``verbalization layer'', i.e., layers that are active exclusively in response to label words in ICL,  resembling the classic example ``grandmother cell'' in neuroscience. We do not follow the traditional view of located representation positing that a specific vehicle of the representation (neurons or small sets of neurons) could be responsible for representing complex concepts. What we find and we will claim is that the model \textit{perform or implement} the high-level \funcverb consistently in certain layers under the background conditions we test (prompting strategies, etc).


\section{Illustrations of the Reconstructed Task Intervention}
\begin{figure}[!htb]
    \centering
    \includegraphics[width=1.0\linewidth]{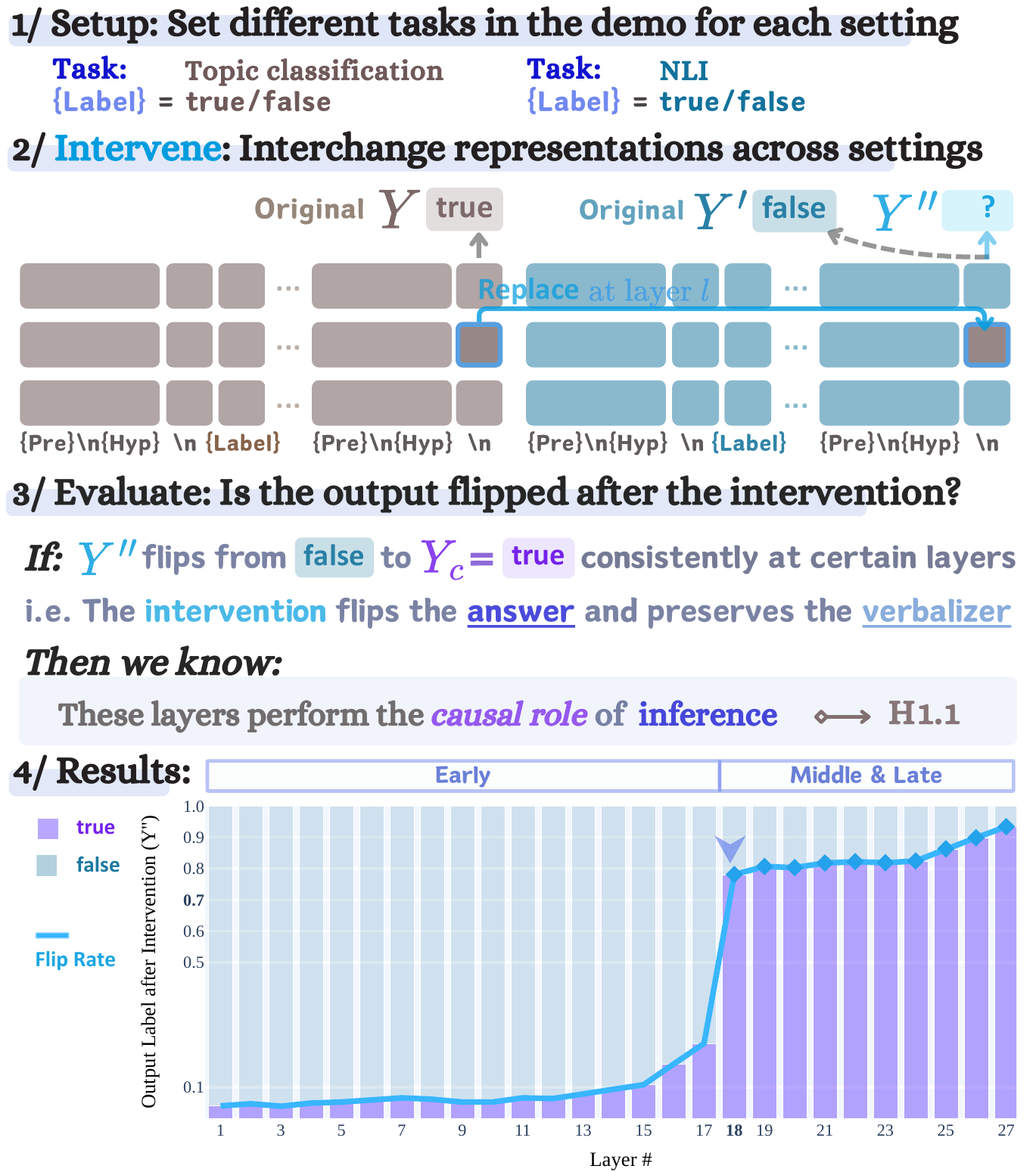}
    \caption{\textbf{Experiment with reconstructed tasks on MultiNLI.} (1) We induce changes in the \coloredinfer by prompting the model to perform a different task on the same input space. (2) We perform intervention by taking the representation from the {\color{task-source}\textbf{sourced model}} (prompted by the alternative task) and replacing the representation of the {\color{task-intervened}\textbf{intervened model}} (prompted by the NLI task) at layer $l$. (3) Intervention effects are evaluated by calculating the matching rate between the intervened output $Y''$ and the hypothetical \cf $Y_c$.}
    \label{fig:illu-change-task}
\end{figure}

Figure~\ref{fig:illu-change-task} illustrates and summarizes our reconstructed task intervention outlined in Section~\ref{sec:change-task}.

\section{Implications of Poor ICL Performance}
\label{interpret-negative-cases}
\begin{figure*}
    \centering
    \includegraphics[width=1\textwidth]{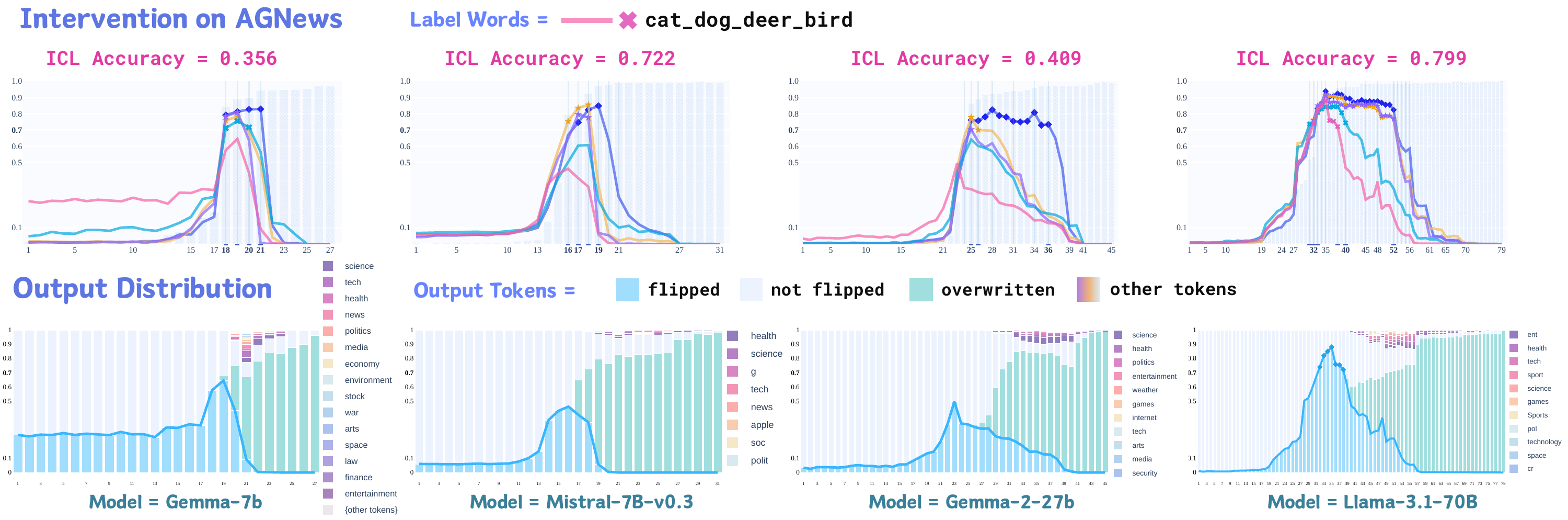}
    \caption{\textbf{Exploration of negative cases where the ICL performance is low.} \textbf{Top:} flip rates of intervention with remapped label spaces. \textbf{Bottom:} Distribution of predicted token from the intervened model with the under-performing label words.
    }
    
    \label{fig:analyses-negative-case-agnews}
\end{figure*}

We do not expect to observe effective intervention results on cases where the ICL performance is inadequate, and such cases will not weaken our argument. This is because the poor ICL performance is very likely to come from the model's failure to develop the necessary abstractions for solving the task, which would include recognizing the task, inferring the correct answer, recognizing the label space, and verbalizing the answer correctly. If the model achieves high ICL performance with some label spaces while not with others, it may employ different sets of abstractions to solve these two ICL tasks.

Nonetheless, we intentionally include some cases where the model fails to solve the ICL task, as complementary results (ICL performance is summarized in Table~\ref{tab:middle_stage} with the low accuracy colored in orange or red). In our experiments, we find that (1) if the ICL accuracy is good enough, e.g., $\geq 0.7$, most settings will show a strong pattern, as summarized before; while some inconsistencies may exist, though rarely. One example is the case where \textsc{Mistral-7B-v0.3} is used on AGNews with the label words with the label words ``cat''/``dog'', as shown in Figure~\ref{fig:analyses-negative-case-agnews}, second column); (2) if the ICL is not good enough (around $0.6$ for binary classification and around $0.4$ for four-class classification), we find the flip rates still generally follow the same trend, though the patterns are \textit{weaker} than settings with high ICL accuracy due to the lack of necessary abstractions developed. Examples include: cases with the label space ``foo''/``bar''  and ``black''/``white'' for \textsc{Gemma-7b} and \textsc{Mistral-7B-v0.3} on RTE (Figure~\ref{fig:results-change-label-nli}, second row) and ``cat''/``dog''/``deer''/``bird'' for \textsc{Mistral-7B-v0.3} on AGNews (Figure~\ref{fig:analyses-negative-case-agnews}, first and third column).

\section{Prompting Strategy to Improve ICL Performance}
\label{prompting-strategy}

We slightly adjust the selection of irrelevant label words and prompting strategies for each setting, to facilitate the model in achieving adequate ICL performance. 

We observe that reproducing the phenomena of interest on \textit{base models} can be challenging with some settings. For instance, 7-8 models on ANLI are difficult to handle with irrelevant label words despite performing adequately with default label words. And AGNews is generally hard for all models because it is a four-class classification and naturally requires more shots.

We adjust our template selection based on the corresponding ICL performance. We always start with a 16-shot demonstration with the ``sentence'' template, aiming for ICL performance above 0.7. If this benchmark is not met, we increase the demonstration to 32 shots. Should challenges persist, we modify the template to ``sent\_label'' and consider adding specific keywords in front of the \texttt{{answer}} to implicitly hint for the intended tasks, such as ``topic:'' for AGNews and ``sentiment:'' for IMDb.


\paragraph{Note on the notion of generalizability.} By the generalizability of our findings, we do not mean that a single set of irrelevant labels should work universally across all models in all settings. Again, our study of causal mechanisms focuses on cases where the model achieves high ICL performance, which means it develops the necessary causal abstractions required to solve the task we want to study. Since LMs are sensitive to contexts and to different prompting strategies, the required settings unsurprisingly vary model by model.

\paragraph{Address concerns about cherry-picking.} Concerns may arise that our selection of successful cases constitutes cherry-picking. 
However, we select cases based on their ICL performance, not their intervention results, and we test a variety of irrelevant label words that are representative of broader cases. We also discuss cases where ICL performance is not good and find that they still generally follow the patterns, though they do so to a lesser extent. (See Appendix~\ref{interpret-negative-cases}).

\section{Predicted Tokens from the experiment with remapped label spaces}
\label{sec:appendix-output-token}

\begin{figure*}
    \centering
    \includegraphics[width=1\textwidth]{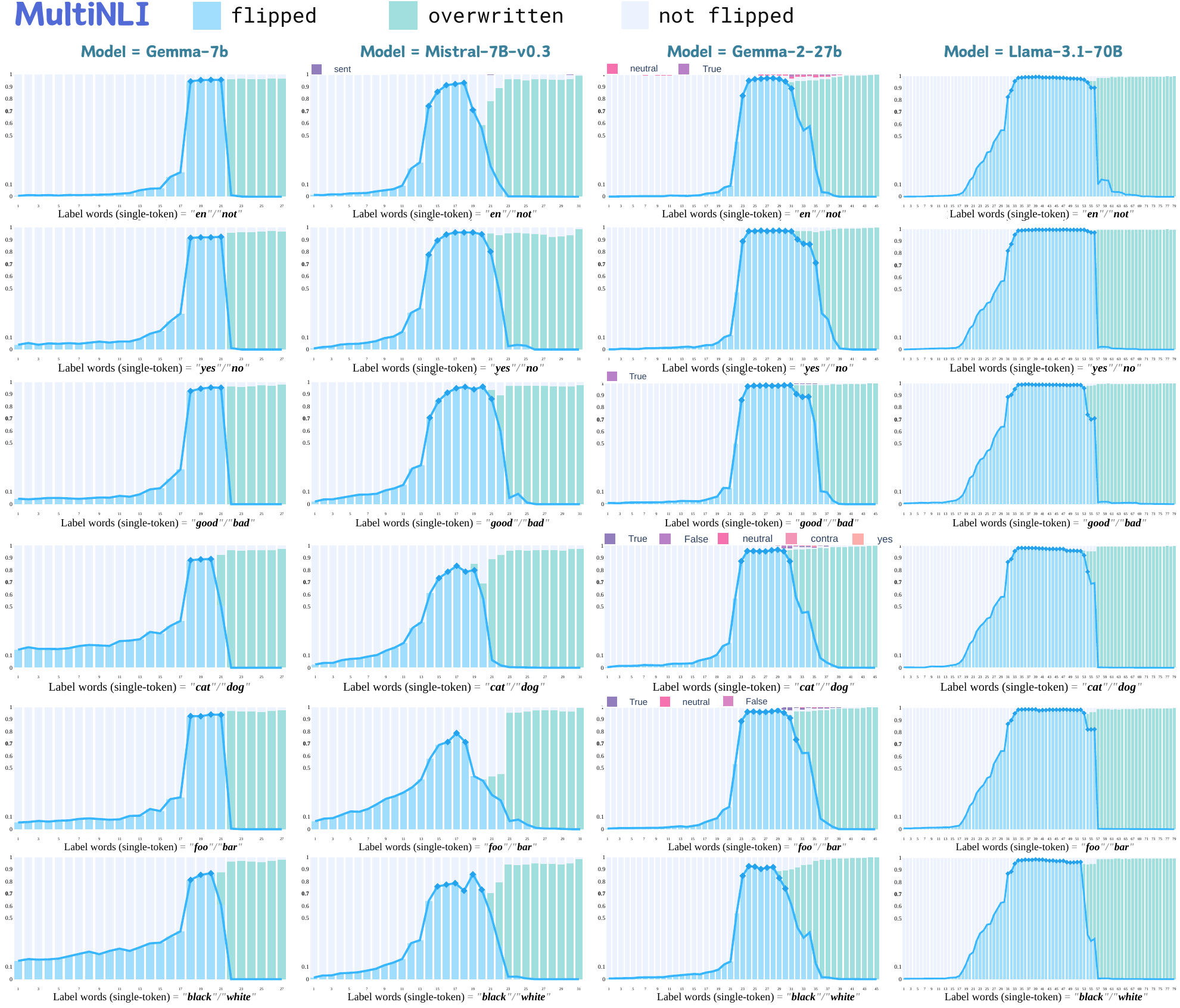}
    \caption{Full results of intervened model's prediction distributions on the MultiNLI dataset.}
    \label{fig:token-multinli}
\end{figure*}

\begin{figure*}
    \centering
    \includegraphics[width=1\textwidth]{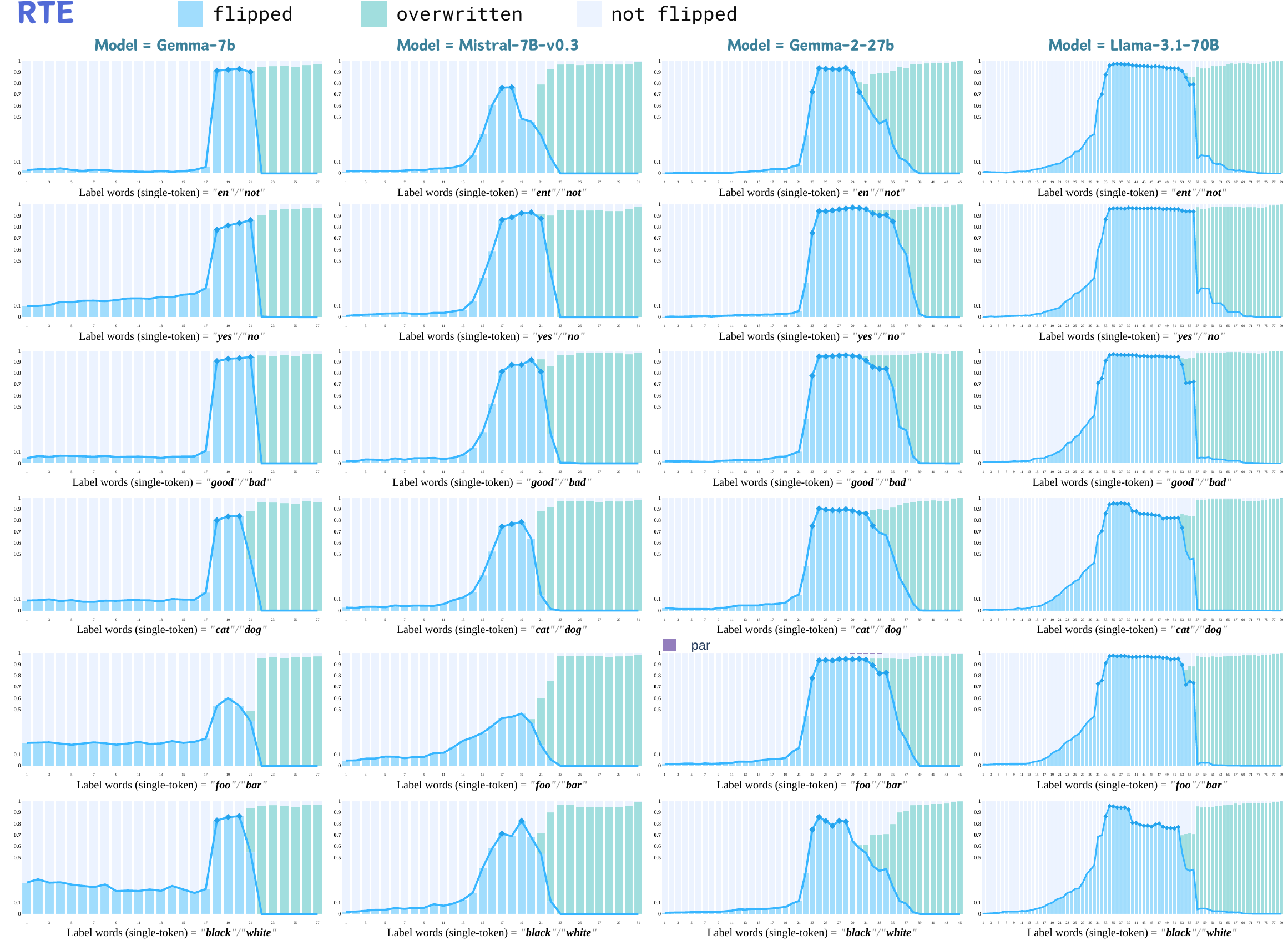}
    \caption{Full results of intervened model's prediction distributions on the RTE dataset.}
    \label{fig:token-rte}
\end{figure*}

\begin{figure*}
    \centering
    \includegraphics[width=1\textwidth]{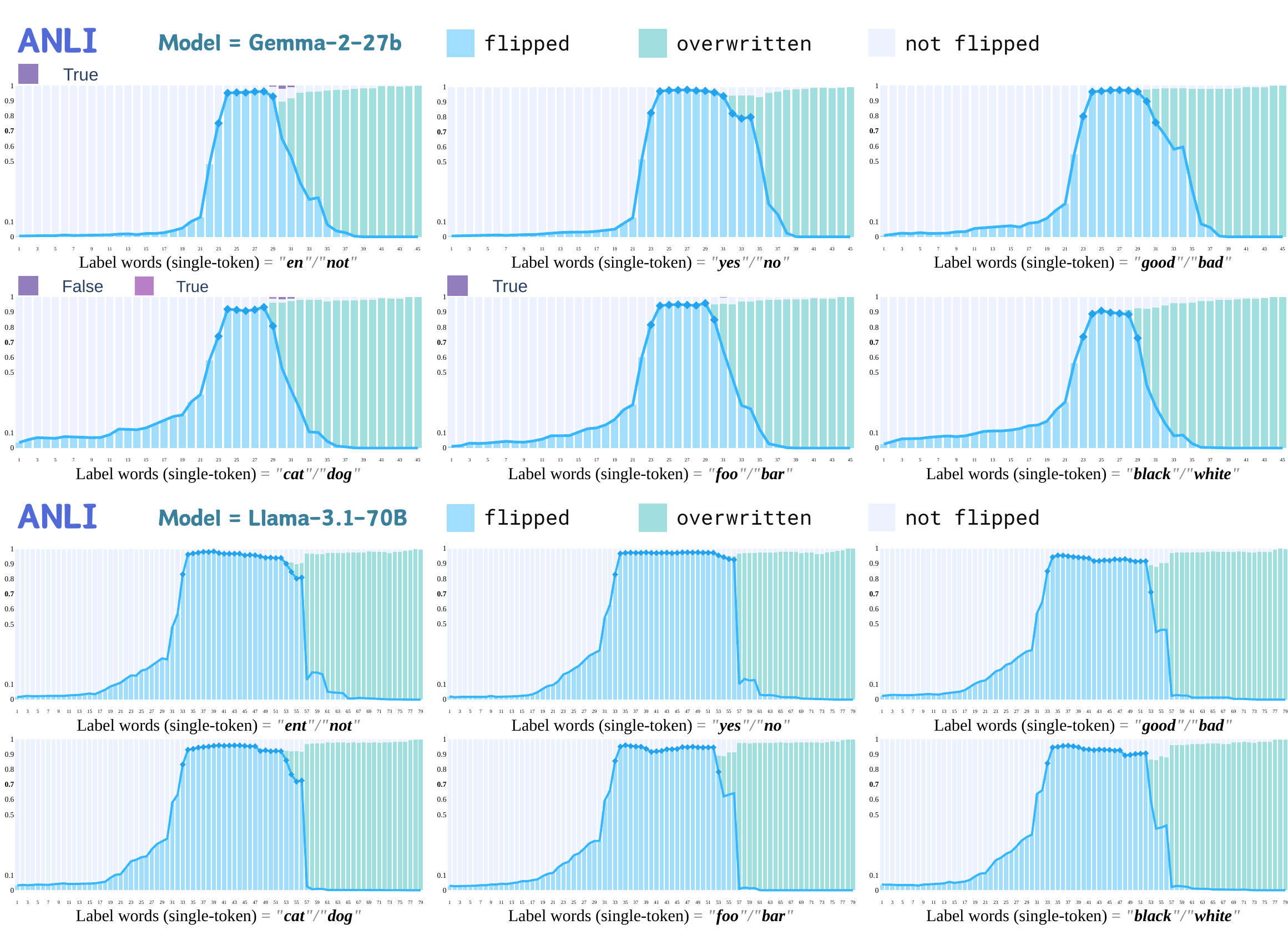}
    \caption{Full results of intervened model's prediction distributions on the ANLI dataset.}
    \label{fig:token-anli}
\end{figure*}

\begin{figure*}
    \centering
    \includegraphics[width=1\textwidth]{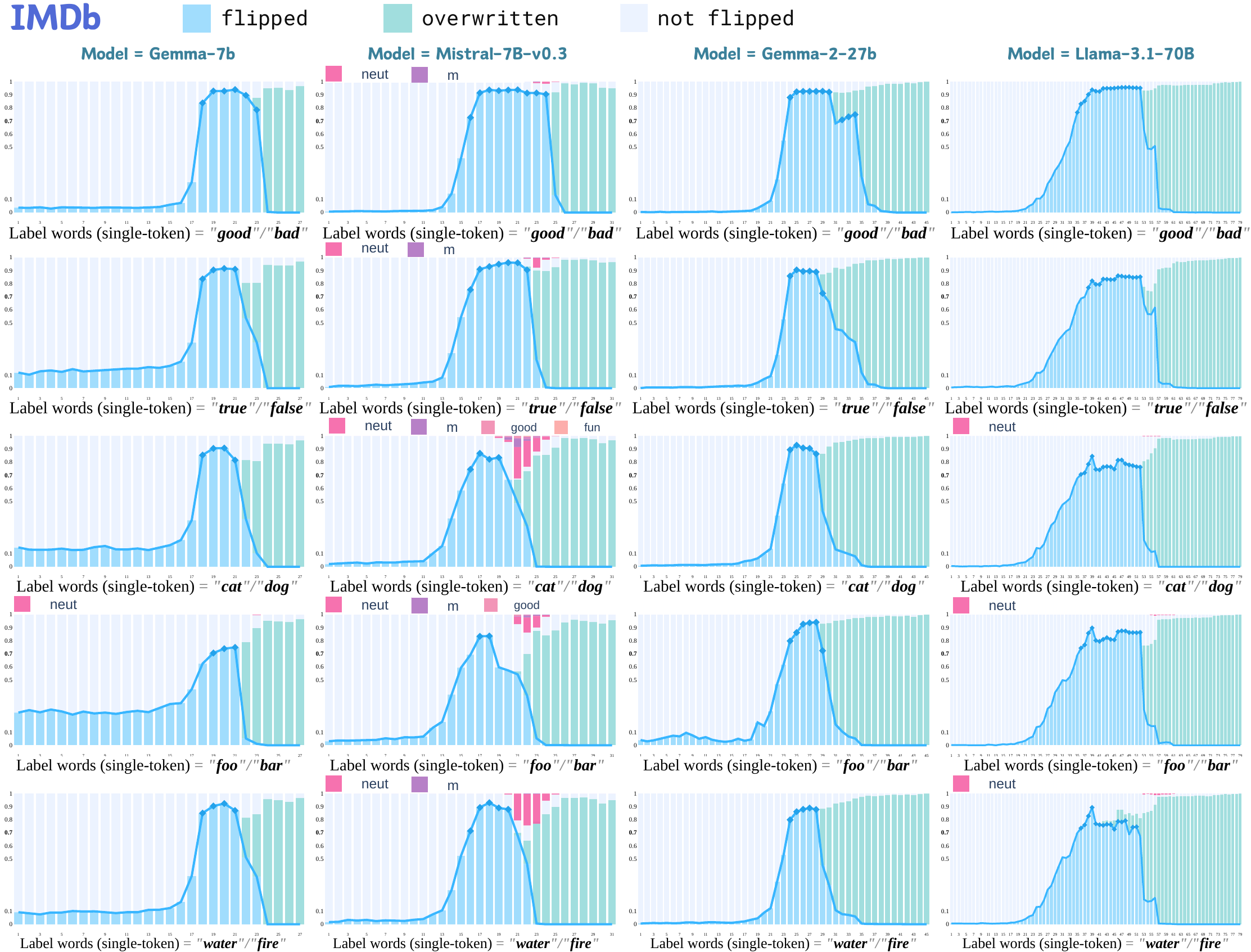}
    \caption{Full results of intervened model's prediction distributions on the IMDb dataset.}
    \label{fig:token-imdb}
\end{figure*}

\begin{figure*}
    \centering
    \includegraphics[width=1\textwidth]{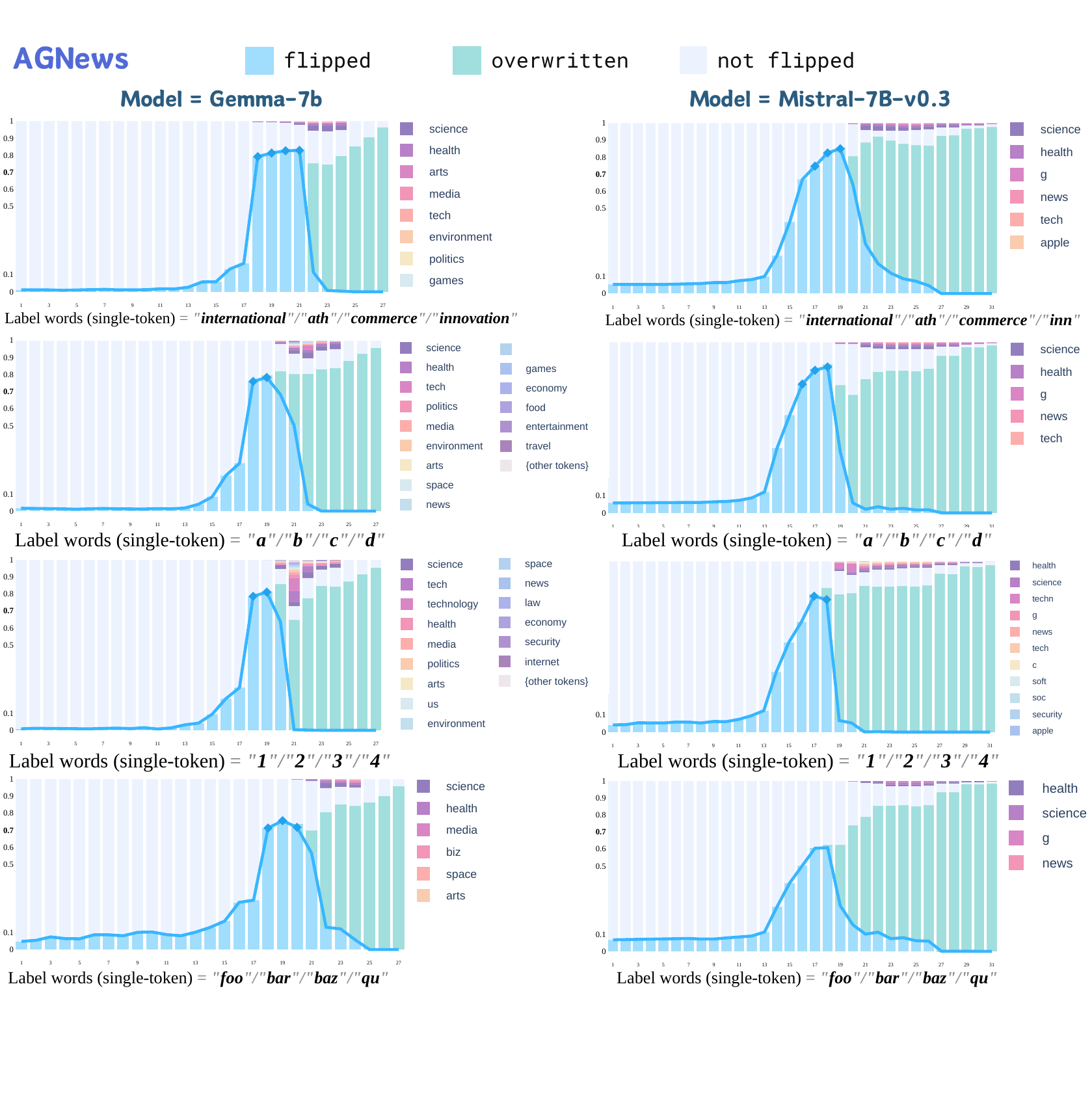}
    \caption{Full results of intervened model's prediction distributions on the AGNews dataset. Part 1.}
    \label{fig:token-agnews1}
\end{figure*}

\begin{figure*}
    \centering
    \includegraphics[width=1\textwidth]{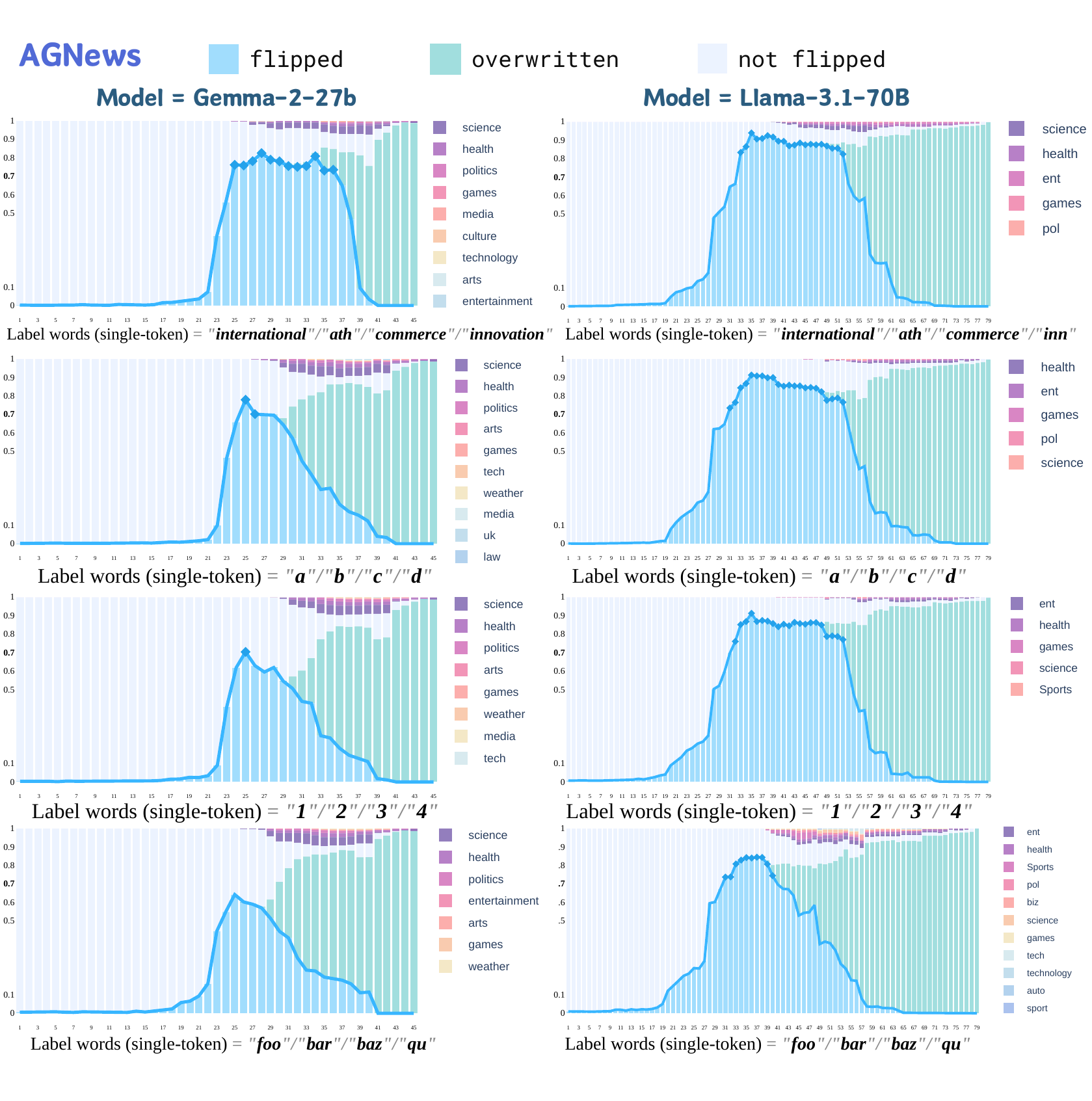}
    \caption{Full results of intervened model's prediction distributions on the AGNews dataset. Part 2.}
    \label{fig:token-agnews2}
\end{figure*}

We present the full results of the intervened model's prediction distributions on all five datasets used in our intervention with remapped label spaces: MultiNLI (Figure~\ref{fig:token-multinli}), RTE (Figure~\ref{fig:token-rte}), ANLI (Figure~\ref{fig:token-anli}), IMDb (Figure~\ref{fig:token-imdb}), AGNews (Figure~\ref{fig:token-agnews1} \&~\ref{fig:token-agnews2}).

\begin{table*}[!htb]
\resizebox{\textwidth}{!}{%
\begin{tabular}{@{}llllll@{}}
\toprule
 &
   &
  \multicolumn{4}{c}{\textbf{Predicted Single Tokens}} \\ \cmidrule(l){3-6} 
\multirow{-2}{*}{\textbf{Dataset}} &
  \multirow{-2}{*}{\textbf{Verbalizers}} &
  \textbf{\textsc{Gemma-7b}} &
  \textbf{\textsc{Mistral-7B-v0.3}} &
  \textbf{\textsc{Gemma-2-27b}} &
  \textbf{\textsc{Llama-3.1-70B}} \\ \midrule
MultiNLI/ &
  \textit{"true"/"false"} &
  \textit{"true"/"false"} &
  \textit{"true"/"false"} &
  \textit{"true"/"false"} &
  \textit{"true"/"false"} \\
RTE/ &
  "en"/"not" &
  "en"/"not" &
  "ent"/"not" &
  "en"/"not" &
  "ent"/"not" \\
ANLI &
  "yes"/"no" &
  "yes"/"no" &
  "yes"/"no" &
  "yes"/"no" &
  "yes"/"no" \\
 &
  "good"/"bad" &
  "good"/"bad" &
  "good"/"bad" &
  "good"/"bad" &
  "good"/"bad" \\
 &
  "cat"/"dog" &
  "cat"/"dog" &
  "cat"/"dog" &
  "cat"/"dog" &
  "cat"/"dog" \\
 &
  "foo"/"bar &
  "foo"/"bar &
  "foo"/"bar &
  "foo"/"bar &
  "foo"/"bar \\
 &
  "black"/"white" &
  "black"/"white" &
  "black"/"white" &
  "black"/"white" &
  "black"/"white" \\ \midrule
IMDb &
  \textit{"positive"/"negative"} &
  {\textit{"positive"/"negative"}} &
  {\textit{"pos"/"negative"}} &
  \textit{"positive"/"negative"} &
  \textit{"positive"/"negative"} \\
 &
  "good"/"bad" &
  "good"/"bad" &
  "good"/"bad" &
  "good"/"bad" &
  "good"/"bad" \\
 &
  "true"/"false" &
  "true"/"false" &
  "true"/"false" &
  "true"/"false" &
  "true"/"false" \\
 &
  "cat"/"dog" &
  "cat"/"dog" &
  "cat"/"dog" &
  "cat"/"dog" &
  "cat"/"dog" \\
 &
  "foo"/"bar" &
  "foo"/"bar" &
  "foo"/"bar" &
  "foo"/"bar" &
  "foo"/"bar" \\
 &
  "water"/"fire" &
  "water"/"fire" &
  "water"/"fire" &
  "water"/"fire" &
  "water"/"fire" \\ \midrule
AGNews &
  \textit{\begin{tabular}[c]{@{}l@{}}"world"/"sports"/\\ "business"/"sci/tech"\end{tabular}} &
  \textit{\begin{tabular}[c]{@{}l@{}}"world"/"sports"/\\ "business"/"sci"\end{tabular}} &
  \textit{\begin{tabular}[c]{@{}l@{}}"world"/"s"/\\ "business"/"sc"\end{tabular}} &
  \textit{\begin{tabular}[c]{@{}l@{}}"world"/"sports"/\\ "business"/"sci"\end{tabular}} &
  \textit{\begin{tabular}[c]{@{}l@{}}"world"/"sports"/\\ "business"/"sci"\end{tabular}} \\
 &
  \begin{tabular}[c]{@{}l@{}}"international"/"athletics"/\\ "commerce"/"innovation"\end{tabular} &
  \begin{tabular}[c]{@{}l@{}}"international"/"ath"/\\ "commerce"/"innovation"\end{tabular} &
  \begin{tabular}[c]{@{}l@{}}"intern"/"ath"/\\ "commerce"/"inn"\end{tabular} &
  \begin{tabular}[c]{@{}l@{}}"international"/"ath"/\\ "commerce"/"innovation"\end{tabular} &
  \begin{tabular}[c]{@{}l@{}}"international"/"ath"/\\ "commerce"/"inn"\end{tabular} \\
 &
  "a"/"b"/"c"/"d" &
  "a"/"b"/"c"/"d" &
  "a"/"b"/"c"/"d" &
  "a"/"b"/"c"/"d" &
  "a"/"b"/"c"/"d" \\
 &
  "1"/"2"/"3"/"4" &
  "1"/"2"/"3"/"4" &
  "1"/"2"/"3"/"4" &
  "1"/"2"/"3"/"4" &
  "1"/"2"/"3"/"4" \\
 &
  "foo"/"bar"/"baz"/"qux" &
  "foo"/"bar"/"baz"/"qu" &
  "foo"/"bar"/"b"/"qu" &
  "foo"/"bar"/"baz"/"qu" &
  "foo"/"bar"/"baz"/"qu" \\ \bottomrule
\end{tabular}%
}
\caption{\textbf{Tokenization of verbalizers}: LM tokenizers may break a word into sub-tokens. Our intervention method requires verbalizers to be differentiable by the first sub-token since it only evaluates the first generated token. Here we present how each model's tokenizer break-down the verbalizers.} 
\label{tab:results-single-tokens}
\end{table*}

\begin{table*}[!htb]
\centering
\resizebox{0.8\textwidth}{!}{%
\begin{tabular}{lllccc}
\toprule
\multirow{2}{*}{\textbf{Intervention}} & \multirow{2}{*}{\textbf{Task}} & \multirow{2}{*}{\textbf{Dataset}} & \multicolumn{3}{c}{\textbf{\# of Test Examples}} \\
             &                      &          & \textbf{Original} & \textbf{Experiments (7-8b models)} & \textbf{Experiments (27-70b models)} \\ \midrule
Change Verbalizer & Natural              & RTE      & 277     & 277         & 277       \\
             & Language             & MultiNLI & 9820    & 300         & 300      \\
             & Inference            & ANLI     & 1000    & 300         & 300      \\ 
             & Sentiment Analysis   & IMDb     & 25000   & 300         & 300        \\ 
             & Topic Classification & AGNews   & 7600    & 300         & 300        \\ \midrule
Change Task  & Multiple Tasks       & MultiNLI & 9820    & 300         & 300     \\ \bottomrule
\end{tabular}%
}
\caption{\textbf{Statistic of datasets:} Due to the computational budget, we reduced the amounts of test examples for all datasets other than the RTE to around 300 examples which we believe is sufficient to test the generalizability of our hypothesized functions. Subsets are sampled randomly using a fix random seed 42 from each corresponding task dataset.}
\label{tab:dataset_stat}
\end{table*}

\begin{table*}[!htb]
\resizebox{\textwidth}{!}{%
\begin{tabular}{@{}lllcccccccc@{}}
\toprule
{} &
  {} &
  {} &
  \multicolumn{2}{c}{{\textbf{\textsc{Gemma-7b}}}} &
  \multicolumn{2}{c}{{\textbf{\textsc{Mistral-7B-v0.3}}}} &
  \multicolumn{2}{c}{{\textbf{\textsc{Gemma-2-27b}}}} &
  \multicolumn{2}{c}{{\textbf{\textsc{Llama-3.1-70B}}}} \\ \cmidrule(l){4-11} 
\multirow{-2}{*}{{\textbf{Intervention}}} &
  \multirow{-2}{*}{{\textbf{Task}}} &
  \multirow{-2}{*}{{\textbf{Dataset}}} &
  {\textbf{Shot}} &
  {\textbf{Template}} &
  {\textbf{Shot}} &
  {\textbf{Template}} &
  {\textbf{Shot}} &
  {\textbf{Template}} &
  {\textbf{Shot}} &
  {\textbf{Template}} \\ \midrule
{Change Verbalizer} &
  {Natural} &
  {MultiNLI} &
  {16} &
  {sentence} &
  {16} &
  {sentence} &
  {16} &
  {sentence} &
  {16} &
  {sentence} \\ \cmidrule(l){3-11} 
{} &
  {Language} &
  {RTE} &
  {16} &
  {sent\_label} &
  {16} &
  {sent\_label} &
  {16} &
  {sentence} &
  {16} &
  {sentence} \\
{} &
  {Inference} &
  {ANLI} &
  {/} &
  {/} &
  {/} &
  {/} &
  {16} &
  {sentence} &
  {16} &
  {sentence} \\ \cmidrule(l){2-11} 
{} &
  {Sentiment Analysis} &
  {IMDb} &
  {16} &
  {passage\_label} &
  {16} &
  {passage\_label} &
  {8} &
  {passage\_label} &
  {16} &
  {passage\_sentiment} \\ \cmidrule(l){2-11} 
{} &
  {Topic Classification} &
  {AGNews} &
  {32} &
  {text\_topic} &
  {32} &
  {text\_topic} &
  {32} &
  {text\_topic} &
  {32} &
  {text\_linebreak} \\ \midrule
{Change Task} &
  {Multiple Tasks} &
  {MultiNLI} &
  {32} &
  {ambi\_instruct} &
  {32} &
  {ambi\_instruct} &
  {16} &
  {ambi\_instruct} &
  {16} &
  {ambi\_instruct} \\ \bottomrule
\end{tabular}%
}
\caption{\textbf{ICL settings details}: we select the combinations of the prompt template and number of demonstrations based on the \textbf{ICL performance}. We search the optimal number of shots in $\{8, 16, 24, 32\}$. See Table~\ref{tab:prompt} for details of prompt templates and Table~\ref{tab:ambi-instruction} for instructions used in the Change Task setting.}
\label{tab:icL_setting}
\end{table*}


\begin{table*}[!htb]
  \centering
  \resizebox{\textwidth}{!}{%
  \begin{tabular}{@{}lllcccccc@{}}
  \toprule
   &
     &
     &
    \multicolumn{3}{c}{\textbf{\textsc{Gemma-7b}}} &
    \multicolumn{3}{c}{\textbf{\textsc{Mistral-7B-v0.3}}} \\ \cmidrule(l){4-9} 
  \multirow{-2}{*}{\textbf{Dataset}} &
    \multirow{-2}{*}{\textbf{\begin{tabular}[c]{@{}l@{}}Intervention\\ on Label Words\end{tabular}}} &
    \multirow{-2}{*}{\textbf{Label Words}} &
    \textbf{sentence 16shots}&
    \textbf{sentence 32shots}&
    {\color[HTML]{3166FF} \textbf{sent\_label 16shots}}&
    \textbf{sentence 16shots}&
    \textbf{sentence 32shots}&
    {\color[HTML]{3166FF} \textbf{sent\_label 16shots}}\\ \midrule
  \textbf{RTE} &
    \textit{\textbf{default → default}} &
    \textit{\textbf{"true"/"false"}} &
    {\color[HTML]{333333} \textit{\textbf{0.807}}} &
    \textit{\textbf{0.786}} &
    {\color[HTML]{333333} \textit{\textbf{0.819}}} &
    \textit{\textbf{0.827}} &
    \textit{\textbf{0.838}} &
    \textit{\textbf{0.813}} \\
   &
    default → relevant &
    "en"/"not" &
    {\color[HTML]{333333} 0.798} &
    {\color[HTML]{333333} 0.787} &
    {\color[HTML]{333333} 0.788} &
    0.755 &
    {\color[HTML]{333333} 0.795} &
    0.773 \\
   &
    default → irrelevant &
    "yes"/"no" &
    {\color[HTML]{333333} 0.838} &
    {\color[HTML]{333333} 0.795} &
    {\color[HTML]{333333} 0.745} &
    0.742 &
    {\color[HTML]{333333} 0.826} &
    0.771 \\
  \textbf{} &
     &
    "good"/"bad" &
    0.780 &
    0.767 &
    0.810 &
    0.721 &
    0.767 &
    0.724 \\
  \textbf{} &
     &
    "cat"/"dog" &
    {\color[HTML]{F8A102} 0.574} &
    {\color[HTML]{9A0000} 0.621} &
    0.759 &
    {\color[HTML]{F8A102} 0.558} &
    {\color[HTML]{F8A102} 0.573} &
    {\color[HTML]{9A0000} 0.637} \\
  \textbf{} &
     &
    "foo"/"bar &
    {\color[HTML]{333333} 0.777} &
    {\color[HTML]{333333} 0.761} &
    {\color[HTML]{9A0000} 0.633} &
    {\color[HTML]{F8A102} 0.536} &
    {\color[HTML]{F8A102} 0.539} &
    {\color[HTML]{F8A102} 0.567} \\
  \textbf{} &
     &
    "black"/"white" &
    {\color[HTML]{9A0000} 0.611} &
    {\color[HTML]{F8A102} 0.537} &
    {\color[HTML]{9A0000} 0.620} &
    {\color[HTML]{9A0000} 0.607} &
    {\color[HTML]{9A0000} 0.639} &
    {\color[HTML]{9A0000} 0.609} \\ \bottomrule
  \end{tabular}%
  }
  \\ 
  \vspace{0.8em}
  \centering
  \resizebox{0.7\textwidth}{!}{%
  \begin{tabular}{@{}lllcccc@{}}
  \toprule
            &  &                 & \multicolumn{2}{c}{\textbf{\textsc{Gemma-2-27b}}}                    & \multicolumn{2}{c}{\textbf{\textsc{Llama-3.1-70B}}} \\ \cmidrule(l){4-7} 
  \multirow{-2}{*}{\textbf{Dataset}} &
    \multirow{-2}{*}{\textbf{\begin{tabular}[c]{@{}l@{}}Intervention\\ on Label Words\end{tabular}}} &
    \multirow{-2}{*}{\textbf{Label Words}} &
    \textbf{sent\_label 8shots}&
    \textbf{\color[HTML]{3166FF} sent\_label 16shots}&
    \textbf{sentence 8shots} &
    \textbf{\color[HTML]{3166FF} sentence 16shots} \\ \midrule
  \textbf{ANLI} &
    \textit{\textbf{default → default}} &
    \textit{\textbf{"true"/"false"}} &
    \textit{\textbf{0.824}} &
    \textit{\textbf{0.851}} &
    \textit{\textbf{0.840}}&
    \textit{\textbf{0.854}} \\
            & default → relevant    & "en"/"not"      & 0.801                        & 0.816                        & 0.713& 0.841  \\
            & default → irrelevant  & "yes"/"no"      & 0.811                        & 0.843                        & 0.731& 0.821  \\
            & \textit{}             & "good"/"bad"    & 0.758                        & 0.790                        & 0.701& 0.802  \\
  \textbf{} &                       & "cat"/"dog"     & {\color[HTML]{9A0000} 0.628} & {\color[HTML]{9A0000} 0.643} &                                   0.731& 0.766  \\
            &                       & "foo"/"bar      & {\color[HTML]{9A0000} 0.633} & 0.728                        & 0.699& 0.808  \\
            &                       & "black"/"white" & {\color[HTML]{9A0000} 0.667} & 0.710                        &                                   0.706& 0.794  \\ \bottomrule
  \end{tabular}%
  }
  \caption{Demonstration of prompt selection. We show results on RTE for \textsc{Gemma-7b} and \textsc{Mistral-7B-v0.3}, and on ANLI for \textsc{Gemma-2-27b} and \textsc{Llama-3.1-70B}, to illustrate the rationale behind our final choice of experimental settings. Behavioral testing typically begins with the "sentence" template (with or without the "label" suffix), using 16-shot prompts for 7b models and 8-shot prompts for larger models. Selected settings are highlighted in bold and colored in blue. \textsc{Llama-3.1-70B} in general does not work well with the "label" suffix.}
  \label{tab:setting-justification}
  \end{table*}

\begin{table*}[!htb]
\resizebox{\textwidth}{!}{%
\begin{tabular}{@{}lllcccccccccccc@{}}
\toprule
 &
   &
   &
  \multicolumn{3}{c}{\textbf{\textsc{Gemma-7b}}} &
  \multicolumn{3}{c}{\textbf{\textsc{Mistral-7B-v0.3}}} &
  \multicolumn{3}{c}{\textbf{\textsc{Gemma-2-27b}}} &
  \multicolumn{3}{c}{\textbf{\textsc{Llama-3.1-70B}}} \\ \cmidrule(l){4-15} 
\multirow{-2}{*}{\textbf{Dataset}} &
  \multirow{-2}{*}{\textbf{\begin{tabular}[c]{@{}l@{}}Intervention \\ on Label Words\end{tabular}}} &
  \multirow{-2}{*}{\textbf{Label Words}} &
  \textbf{ICL} &
  \textbf{Start} &
  \textbf{End} &
  \textbf{ICL} &
  \textbf{Start} &
  \textbf{End} &
  \textbf{ICL} &
  \textbf{Start} &
  \textbf{End} &
  \textbf{ICL} &
  \textbf{Start} &
  \textbf{End} \\ \midrule
\textbf{MultiNLI} &
  \textit{\textbf{default → default}} &
  \textit{\textbf{"true"/"false"}} &
  {\color[HTML]{000000} \textit{\textbf{0.889}}} &
  \textit{\textbf{18}} &
  \textit{\textbf{27}} &
  \textit{\textbf{0.842}} &
  \textit{\textbf{15}} &
  \textit{\textbf{31}} &
  \textit{\textbf{0.939}} &
  \textit{\textbf{24}} &
  \textit{\textbf{45}} &
  \textit{\textbf{0.956}} &
  \textit{\textbf{34}} &
  \textit{\textbf{79}} \\
 &
  default → relevant &
  "en"/"not" &
  {\color[HTML]{000000} 0.921} &
  18 &
  21 &
  0.870 &
  15 &
  18 &
  0.943 &
  24 &
  29 &
  0.958 &
  34 &
  52 \\
\textbf{} &
   &
  "yes"/"no" &
  {\color[HTML]{000000} 0.856} &
  18 &
  21 &
  0.820 &
  15 &
  20 &
  0.944 &
  24 &
  31 &
  0.959 &
  34 &
  53 \\
 &
   &
  "good"/"bad" &
  {\color[HTML]{000000} 0.834} &
  18 &
  21 &
  0.747 &
  15 &
  20 &
  0.913 &
  24 &
  31 &
  0.950 &
  34 &
  52 \\
\textbf{} &
  default → irrelevant &
  "cat"/"dog" &
  {\color[HTML]{000000} 0.702} &
  18 &
  20 &
  {\color[HTML]{000000} 0.691} &
  15 &
  19 &
  0.912 &
  24 &
  29 &
  0.949 &
  34 &
  52 \\
\textbf{} &
   &
  "foo"/"bar &
  {\color[HTML]{000000} 0.770} &
  18 &
  21 &
  {\color[HTML]{9A0000} 0.616} &
  15 &
  17 &
  0.908 &
  24 &
  29 &
  0.950 &
  34 &
  52 \\
\textbf{} &
   &
  "black"/"white" &
  {\color[HTML]{000000} 0.701} &
  18 &
  20 &
  {\color[HTML]{000000} 0.772} &
  15 &
  19 &
  0.906 &
  24 &
  28 &
  0.944 &
  34 &
  52 \\ \midrule
\textbf{RTE} &
  \textit{\textbf{default → default}} &
  \textit{\textbf{"true"/"false"}} &
  {\color[HTML]{333333} \textit{\textbf{0.819}}} &
  \textit{\textbf{18}} &
  \textit{\textbf{27}} &
  \textit{\textbf{0.813}} &
  \textit{\textbf{17}} &
  \textit{\textbf{31}} &
  \textit{\textbf{0.857}} &
  \textit{\textbf{24}} &
  \textit{\textbf{45}} &
  \textit{\textbf{0.854}} &
  \textit{\textbf{34}} &
  \textit{\textbf{79}} \\
 &
  default → relevant &
  "en"/"not" &
  {\color[HTML]{333333} 0.788} &
  {\color[HTML]{333333} 18} &
  21 &
  0.773 &
  {\color[HTML]{333333} 17} &
  18 &
  0.868 &
  {\color[HTML]{333333} 24} &
  28 &
  0.839 &
  {\color[HTML]{333333} 34} &
  52 \\
 &
   &
  "yes"/"no" &
  {\color[HTML]{333333} 0.745} &
  {\color[HTML]{333333} 18} &
  21 &
  0.771 &
  {\color[HTML]{333333} 17} &
  20 &
  0.863 &
  {\color[HTML]{333333} 24} &
  31 &
  0.854 &
  {\color[HTML]{333333} 34} &
  52 \\
\textbf{} &
   &
  "good"/"bad" &
  0.810 &
  18 &
  21 &
  0.724 &
  17 &
  20 &
  0.846 &
  24 &
  30 &
  0.829 &
  34 &
  52 \\
\textbf{} &
  default → irrelevant &
  "cat"/"dog" &
  0.759 &
  18 &
  20 &
  {\color[HTML]{9A0000} 0.637} &
  17 &
  19 &
  0.805 &
  24 &
  31 &
  0.835 &
  34 &
  39 \\
\textbf{} &
   &
  "foo"/"bar &
  {\color[HTML]{9A0000} 0.633} &
  / &
  / &
  {\color[HTML]{F8A102} 0.567} &
  / &
  / &
  0.841 &
  24 &
  31 &
  0.830 &
  34 &
  52 \\
\textbf{} &
   &
  "black"/"white" &
  {\color[HTML]{F8A102} 0.620} &
  18 &
  20 &
  {\color[HTML]{F8A102} 0.609} &
  17 &
  19 &
  0.833 &
  24 &
  28 &
  0.838 &
  34 &
  39 \\ \midrule
\textbf{ANLI} &
  \textit{\textbf{default → default}} &
  \textit{\textbf{"true"/"false"}} &
  \textit{\textbf{0.723}} &
  {\color[HTML]{000000} -} &
  {\color[HTML]{000000} -} &
  {\color[HTML]{9A0000} \textit{\textbf{0.679}}} &
  - &
  - &
  \textit{\textbf{0.851}} &
  \textit{\textbf{24}} &
  \textit{\textbf{45}} &
  \textit{\textbf{0.854}} &
  \textit{\textbf{34}} &
  \textit{\textbf{79}} \\
 &
  default → relevant &
  "en"/"not" &
  {\color[HTML]{9A0000} 0.693} &
  {\color[HTML]{000000} -} &
  {\color[HTML]{000000} -} &
  {\color[HTML]{F8A102} 0.589} &
  - &
  - &
  0.816 &
  24 &
  28 &
  0.841 &
  34 &
  52 \\
 &
   &
  "yes"/"no" &
  {\color[HTML]{9A0000} 0.721} &
  {\color[HTML]{000000} -} &
  {\color[HTML]{000000} -} &
  {\color[HTML]{9A0000} 0.659} &
  - &
  - &
  0.843 &
  24 &
  30 &
  0.821 &
  34 &
  52 \\
 &
  \textit{} &
  "good"/"bad" &
  {\color[HTML]{9A0000} \textit{0.676}} &
  {\color[HTML]{000000} \textit{-}} &
  {\color[HTML]{000000} \textit{-}} &
  {\color[HTML]{9A0000} 0.606} &
  \textit{-} &
  \textit{-} &
  \textit{0.790} &
  24 &
  \textit{29} &
  \textit{0.802} &
  34 &
  \textit{52} \\
\textbf{} &
  default → irrelevant &
  "cat"/"dog" &
  {\color[HTML]{F8A102} 0.522} &
  {\color[HTML]{000000} -} &
  {\color[HTML]{000000} -} &
  {\color[HTML]{F8A102} 0.519} &
  - &
  - &
  {\color[HTML]{9A0000} 0.643} &
  24 &
  28 &
  0.766 &
  34 &
  52 \\
 &
   &
  "foo"/"bar &
  {\color[HTML]{F8A102} 0.601} &
  {\color[HTML]{000000} -} &
  {\color[HTML]{000000} -} &
  {\color[HTML]{F8A102} 0.521} &
  - &
  - &
  0.728 &
  {\color[HTML]{333333} 24} &
  29 &
  0.808 &
  {\color[HTML]{333333} 34} &
  52 \\
 &
   &
  "black"/"white" &
  {\color[HTML]{F8A102} 0.511} &
  {\color[HTML]{000000} -} &
  {\color[HTML]{000000} -} &
  {\color[HTML]{F8A102} 0.542} &
  - &
  - &
  0.710 &
  24 &
  28 &
  0.794 &
  34 &
  52 \\ \midrule
\textbf{IMDb} &
  \textit{\textbf{default → default}} &
  \textit{\textbf{"positive"/"negative"}} &
  \textit{\textbf{0.844}} &
  {\color[HTML]{333333} \textit{\textbf{18}}} &
  \textit{\textbf{27}} &
  \textit{\textbf{0.873}} &
  {\color[HTML]{333333} \textit{\textbf{17}}} &
  \textit{\textbf{31}} &
  \textit{\textbf{0.889}} &
  {\color[HTML]{333333} \textit{\textbf{24}}} &
  \textit{\textbf{45}} &
  \textit{\textbf{0.913}} &
  {\color[HTML]{333333} \textit{\textbf{39}}} &
  \textit{\textbf{79}} \\
 &
  default → relevant &
  "good"/"bad" &
  0.887 &
  {\color[HTML]{333333} 18} &
  21 &
  0.904 &
  17 &
  24 &
  0.919 &
  24 &
  30 &
  0.923 &
  39 &
  52 \\
 &
   &
  "true"/"false" &
  0.733 &
  {\color[HTML]{333333} 18} &
  21 &
  0.772 &
  17 &
  21 &
  0.912 &
  24 &
  28 &
  0.918 &
  39 &
  52 \\
\textbf{} &
  default → irrelevant &
  "cat"/"dog" &
  0.709 &
  18 &
  20 &
  0.778 &
  17 &
  17 &
  0.880 &
  24 &
  27 &
  0.908 &
  39 &
  52 \\
\textbf{} &
   &
  "foo"/"bar" &
  {\color[HTML]{CB0000} 0.617} &
  18 &
  21 &
  0.821 &
  17 &
  18 &
  0.740 &
  24 &
  28 &
  0.903 &
  39 &
  52 \\
\textbf{} &
   &
  "water"/"fire" &
  0.760 &
  18 &
  20 &
  0.816 &
  17 &
  18 &
  0.870 &
  24 &
  28 &
  0.888 &
  39 &
  51 \\ \midrule
\textbf{AGNews} &
  \textit{\textbf{default → default}} &
  \textit{\textbf{\begin{tabular}[c]{@{}l@{}}"world"/"sports"/\\ "business"/"sci/tech"\end{tabular}}} &
  \textit{\textbf{0.822}} &
  {\color[HTML]{333333} \textit{\textbf{18}}} &
  \textit{\textbf{27}} &
  \textit{\textbf{0.852}} &
  {\color[HTML]{333333} \textit{\textbf{16}}} &
  \textit{\textbf{31}} &
  \textit{\textbf{0.850}} &
  {\color[HTML]{333333} \textit{\textbf{25}}} &
  \textit{\textbf{45}} &
  \textit{\textbf{0.874}} &
  {\color[HTML]{333333} \textit{\textbf{35}}} &
  \textit{\textbf{79}} \\
 &
  default → irrelevant &
  \begin{tabular}[c]{@{}l@{}}"international"/"ath-\\ -letics"/"commerce"/\\ "innovation"\end{tabular} &
  0.822 &
  {\color[HTML]{333333} 18} &
  21 &
  0.832 &
  {\color[HTML]{333333} 17} &
  19 &
  0.841 &
  {\color[HTML]{333333} 25} &
  34 &
  0.869 &
  {\color[HTML]{333333} 35} &
  48 \\
\textbf{} &
   &
  "a"/"b"/"c"/"d" &
  0.810 &
  {\color[HTML]{333333} 18} &
  19 &
  0.752 &
  16 &
  18 &
  0.834 &
  25 &
  25 &
  0.856 &
  35 &
  47 \\
\textbf{} &
   &
  "1"/"2"/"3"/"4" &
  0.807 &
  18 &
  19 &
  0.723 &
  17 &
  18 &
  0.830 &
  25 &
  25 &
  0.826 &
  35 &
  47 \\
\textbf{} &
   &
  \begin{tabular}[c]{@{}l@{}}"foo"/"bar"/\\ "baz"/"qux"\end{tabular} &
  {\color[HTML]{F8A102} 0.572} &
  18 &
  19 &
  {\color[HTML]{9A0000} 0.683} &
  / &
  / &
  0.769 &
  / &
  / &
  0.811 &
  35 &
  38 \\ \midrule
\textbf{MultiNLI} &
  \textit{\textbf{\begin{tabular}[c]{@{}l@{}}NLI → NLI classification\\ (entail vs. not entail)\end{tabular}}} &
  \textit{\textbf{"true"/"false"}} &
  \textit{\textbf{0.800}} &
  {\color[HTML]{333333} \textit{\textbf{18}}} &
  \textit{\textbf{27}} &
  \textit{\textbf{0.794}} &
  \textbf{17} &
  \textit{\textbf{31}} &
  \textit{\textbf{0.876}} &
  \textit{\textbf{24}} &
  \textit{\textbf{45}} &
  \textit{\textbf{0.860}} &
  \textit{\textbf{34}} &
  \textit{\textbf{79}} \\
\textbf{} &
  \begin{tabular}[c]{@{}l@{}}NLI → Domain classification\\ (government vs. fiction)\end{tabular} &
  "true"/"false" &
  {\color[HTML]{F8A102} 0.606} &
  {\color[HTML]{333333} 18} &
  20 &
  0.864 &
  16 &
  18 &
  0.744 &
  24 &
  28 &
  0.841 &
  40 &
  52 \\
\textbf{} &
  \begin{tabular}[c]{@{}l@{}}NLI → Domain classification\\ (government vs. telephone)\end{tabular} &
  "true"/"false" &
  0.713 &
  {\color[HTML]{333333} 18} &
  20 &
  0.951 &
  16 &
  18 &
  0.936 &
  24 &
  28 &
  0.987 &
  41 &
  52 \\
\textbf{} &
  \begin{tabular}[c]{@{}l@{}}NLI → Detect negation\\ (negation vs. no negation)\end{tabular} &
  "true"/"false" &
  {\color[HTML]{F8A102} 0.590} &
  18 &
  20 &
  {\color[HTML]{F8A102} 0.591} &
  16 &
  18 &
  0.752 &
  24 &
  26 &
  0.782 &
  34 &
  52 \\
\textbf{} &
  \begin{tabular}[c]{@{}l@{}}NLI → Detect Overlap\\ (overlap vs. non-overlap)\end{tabular} &
  "true"/"false" &
  {\color[HTML]{F8A102} 0.584} &
  18 &
  20 &
  0.776 &
  16 &
  18 &
  {\color[HTML]{9A0000} 0.686} &
  24 &
  28 &
  0.811 &
  34 &
  52 \\ \bottomrule
\end{tabular} 
}
\caption{\textbf{ICL Performance Across Settings and Identification of the ``Platitude'' Stage.} The \textit{default $\gets$ default \textit{baseline}} are italic and bold. For \textbf{remap label space} experiment, we delineate the approximate middle stage (start and end of layers) where we achieve the counterfactual scenario. This stage should see peak flip rates, positioned between two monotonic trends—one converging towards the baseline, the other diverging.
For the \textbf{change task} experiment, we summarize the start and end of the first sub-phase (the first platitude we see), aligning with the middle stage of the \textbf{remap label space} setting.
We do not conduct intervention on settings where the ICL performance is inadequate and note in the corresponding cell with ``-'' (for ANLI) or ``/'' (for label words such as ``foo''/``bar" on RTE). Flip rates below 0.6 are highlighted in orange, and those below 0.7 in red.
}
\label{tab:middle_stage}
\end{table*}

\begin{table*}[!htb]
\resizebox{\textwidth}{!}{%
\begin{tabular}{@{}p{\dimexpr0.2\textwidth}p{\dimexpr0.15\textwidth}p{\dimexpr0.7\textwidth}@{}}
\toprule
\textbf{Name} & \textbf{Type} & \textbf{Prompt Templates} \\ 
\midrule
sentence& Blank & Sentence 1: \{premise\}\textbackslash{}nSentence 2: \{hypothesis\}\textbackslash{}n\{answer\}\\ 

sent\_label& Blank & Sentence 1: \{premise\}\textbackslash{}nSentence 2: \{hypothesis\}\textbackslash{}nLabel:\{answer\}\\ 
\midrule
passage\_label& Blank & Passage: \{premise\}\textbackslash{}nLabel:\{answer\}\\
 passage\_sentiment& Imply&Passage: \{premise\}\textbackslash{}nSentiment\textbackslash{}n\{answer\}\\ 
\midrule
text\_linebreak& Blank& Text: \{premise\}\textbackslash{}n\{answer\}\\

text\_topic& Imply& Text: \{premise\}\textbackslash{}nTopic:\{answer\}\\ 
\bottomrule
\end{tabular}%
}
\caption{\textbf{Prompt templates:} We intentionally avoid using prompts with instructions and explicit hints for the verbalizers. We find that \textsc{Llama-3.1-70B} is very sensitive to the ending token of the prompt and often yields degenerated results when the prompt does not end with a newline token. Therefore we use prompts different from other models in IMDb and AGNews.}
\label{tab:prompt}
\end{table*}

\begin{table*}[!htb]
\resizebox{\textwidth}{!}{%
\begin{tabular}{ll}
\hline
\textbf{Task Name}           & \textbf{Classification Task}                                                            \\ \hline
default (nli)                & Determine whether the first sentence entails the second sentence,                       \\
government files vs. fiction & Determine whether the two sentences are from government files or fiction.               \\
government files vs. telephone recordings &
  Determine whether the two sentences are from government files or telephone recordings. \\
detect overlap &
  Determine whether all words from the second sentence also appear in the first sentence. \\
detect negation              & Determine whether the second sentence includes any negation words ("not", "no", "n't"). \\ \hline
\end{tabular}%
}
\caption{\textbf{Alternative tasks for the change task setting}: We use the dataset from \citet{si-etal-2023-measuring}, which provides a variant of MultiNLI with labels for the alternative tasks.}
\label{tab:ambi-task}
\end{table*}

\begin{table*}[!htb]
\resizebox{\textwidth}{!}{%
\begin{tabular}{@{}p{\dimexpr0.1\textwidth}p{\dimexpr0.2\textwidth}p{\dimexpr0.7\textwidth}@{}}
\toprule
\textbf{Name} &
  \textbf{Task} &
  \textbf{Prompt Templates} \\ 
  \midrule
ambi\_inst &
  default (nli) &
  In this task, you will be presented with a premise sentence (the first sentence) and a hypothesis sentence (the second sentence). Determine whether the premise sentence entails (implies) or does not entail the hypothesis sentence. Please answer with \textbackslash{}"\{pos\_label\}\textbackslash{}" for entailment and \textbackslash{}"\{neg\_label\}\textbackslash{}" for non-entailment. \\
 &
  government files vs. fiction &
  In this task, you will be presented with a premise sentence (the first sentence) and a hypothesis sentence (the second sentence). Determine whether they come from government files or fiction. Please answer with \textbackslash{}"\{pos\_label\}\textbackslash{}" for government and \textbackslash{}"\{neg\_label\}\textbackslash{}" for fiction. \\
 &
  government files vs. telephone recordings&
  In this task, you will be presented with a premise sentence (the first sentence) and a hypothesis sentence (the second sentence). Determine whether they come from government files or telephone. Please answer with \textbackslash{}"\{pos\_label\}\textbackslash{}" for government and \textbackslash{}"\{neg\_label\}\textbackslash{}" for telephone.\\
 &
  detect overlap &
  In this task, you will be presented with a premise sentence (the first sentence) and a hypothesis sentence (the second sentence). Determine whether all words in the second sentence also appear in the first sentence. If so, answer \textbackslash{}"\{pos\_label\}\textbackslash{}"; if not, answer \textbackslash{}"\{neg\_label\}\textbackslash{}". \\
 &
  detect negation &
  In this task, you will be presented with a premise sentence (the first sentence) and a hypothesis sentence (the second sentence). Determine whether there are any negation words in the second sentence (\textbackslash{}"not\textbackslash{}", \textbackslash{}"no\textbackslash{}", \textbackslash{}"n't\textbackslash{}"). Please answer with \textbackslash{}"\{pos\_label\}\textbackslash{}" for not having negations and \textbackslash{}"\{neg\_label\}\textbackslash{}" for having negations. \\ 
\bottomrule
\end{tabular}%
}
\caption{\textbf{Instruction prompts for Change Task experiments:} We use the instructions from \citet{si-etal-2023-measuring}. \{pos\_label\} and \{neg\_label\} are replaced with their corresponding verbalizers.}
\label{tab:ambi-instruction}
\end{table*}

\end{document}